\documentclass[%
 reprint,
 superscriptaddress,
 nofootinbib,
 amsmath,
 amssymb,
 aps,
 prx,
]{revtex4-2}

\usepackage{graphicx}
\usepackage{dcolumn}
\usepackage{bm}
\usepackage{enumitem}
\usepackage{physics}
\usepackage{siunitx}
\usepackage{float}
\usepackage{hyperref}
\usepackage{color}
\usepackage{xcolor}
\usepackage{soul}
\usepackage{algorithm}      
\usepackage{algpseudocode} 
\renewcommand{\selectlanguage}[1]{}

\newcommand{\ud}{{\mathrm{d}}}

\begin{document}

\preprint{APS/123-QED}

\title{Using matrix-product states for time-series machine learning}

\author{Joshua B. Moore}
\email{joshua.moore@sydney.edu.au}
\author{Hugo P. Stackhouse}%
\email{hugo.stackhouse@sydney.edu.au}
\author{Ben D. Fulcher}
\affiliation{%
 School of Physics, The University of Sydney, NSW 2006, Australia
}%
\author{Sahand Mahmoodian}
\email{sahand.mahmoodian@sydney.edu.au}
\affiliation{Institute for Photonics and Optical Sciences (IPOS), School of Physics, The University of Sydney, NSW 2006, Australia}
\affiliation{Centre for Engineered Quantum Systems, School of Physics, The University of Sydney, NSW, 2006, Australia}

\date{\today}

\begin{abstract}
Matrix-product states (MPS) have proven to be a versatile ansatz for modeling quantum many-body physics.
For many applications, and particularly in one-dimension, they capture relevant quantum correlations in many-body wavefunctions while remaining tractable (polynomial scaling) to store and manipulate on a classical computer.
This has motivated researchers to also apply the MPS ansatz to machine learning (ML) problems where capturing complex correlations in datasets is also a key requirement.
Here, for the first time, we develop an MPS-based algorithm, MPSTime, for learning a joint probability distribution underlying a time-series dataset, and show how it can be used to tackle important time-series ML problems, including classification and imputation.
MPSTime can efficiently learn complicated time-series probability distributions directly from data, requires only moderate maximum MPS bond dimension $\chi_{\rm{max}}$ (with values for our applications ranging between $\chi_{\rm{max}} = 20-160$), and can be trained for both classification and imputation tasks under a single logarithmic loss function.
Using synthetic and publicly available real-world datasets---spanning applications in medicine, energy, and astronomy---we demonstrate performance competitive with state-of-the-art ML approaches, but with the key advantage of encoding the full joint probability distribution learned from the data, which is useful for analyzing and interpreting its underlying structure.
This manuscript is supplemented with the release of a publicly available code package \textit{MPSTime} that implements our approach, and can be used to reproduce all presented results.
The effectiveness of the MPS-based ansatz for capturing complex correlation structures in time-series data makes it a powerful foundation for tackling challenging time-series analysis problems across science, industry, and medicine.
\end{abstract}

\maketitle


\section{\label{sec:Intro}Introduction}

Over the last few decades tensor network (TN) methods have become widely used tools to simulate quantum many-body systems \cite{Fannes1992COMMP, cirac_matrix_2021, orus_practical_2014, schollwoeck_density-matrix_2011}.
Their success lies in their ability to truncate an exponentially large Hilbert space into a small subspace containing the correlations relevant to the system being studied. 
This means that they allow a balancing of expressiveness and storage.
Tensor network approaches have been used to calculate energy eigenstates and dynamics of condensed matter \cite{verstraete_matrix_2006, Pirvu2012PRB} and quantum optics \cite{manzoni_simulating_2017, mahmoodian_dynamics_2020, Pichler2016PRL, Mahmoodian2019PRL} systems and to simulate quantum computing problems \cite{vidal_efficient_2003, tindall_efficient_2024}. 
A particularly successful and widely used class of TNs are one-dimensional matrix-product states (MPS) also known as tensor trains \cite{Oseledets2011SIAMJSCICOMP}.
Matrix-product states can be used to represent ground states of a large class of one-dimensional systems with short-range interactions. 
Efficient techniques are readily available to calculate observables, correlation functions and to contract MPSs in general \cite{schollwoeck_density-matrix_2011}.
On top of this, significant research into MPS has led to the development of efficient algorithms for finding ground states and time evolution \cite{vidal_efficient_2004}.

The ability of MPS to express a range of correlation functions while remaining computationally tractable makes it an ideal foundation for developing machine-learning (ML) algorithms.
We use the term `machine learning' to describe a class of statistical algorithms designed to solve data-driven problems, such as classification, by learning generalizable patterns and structures from data.
The use of MPS for ML was first realized by \citet{stoudenmireSupervisedLearningQuantumInspired2017}, who showed that MPS can be used to classify numerical digits from (two-dimensional) images of human handwriting.
More specifically, the wavefunction of a spin-$1/2$ chain stored using an MPS can be used to classify handwritten digits in the MNIST dataset with an error $<1\%$.
In this approach real-valued data is encoded into an exponentially large Hilbert space using a nonlinear encoding function.
A classifier is then trained by finding an MPS 
for each class (i.e., digit) having large overlap with data from that class, while being orthogonal with other classes.
This approach was subsequently extended to perform unsupervised generative ML with the MNIST handwriting image classification dataset \cite{Han2018:UnsupervisedGenerativeModeling, Dymarsky2022:TensorNetworkLearn, Cheng2019:TreeTensorNetworks, Ran2020:TensorNetworkCompressed}, and others \cite{Meiburg2023:GenerativeLearningContinuous, Mossi2024:MatrixProductState}.

More recently, MPS-based generative models have been applied to tackle a diverse range of important unsupervised ML problems such as anomaly detection \cite{Wang2020:AnomalyDetectionTensor, Zunkovic2023:PositiveUnlabeledLearning, Aizpurua2024:TensorNetworksExplainable}, image segmentation \cite{Selvan2020:TensorNetworksMedical}, and clustering \cite{Shi2022:ClusteringUsingMatrix}, among others \cite{Bai2022:UnsupervisedRecognitionInformative, Liu2021:EntanglementBasedFeatureExtraction}.
Beyond conventional ML applications, MPS-based generative models have also been applied to the problem of compressed sensing \cite{Ran2020:TensorNetworkCompressed}, where they have demonstrated strong performance in sparse signal reconstruction. 
In the context of supervised ML, MPS-based classifiers have been applied to a multitude of real-world datasets including weather data \cite{bhatia_matrix_2019}, medical images \cite{Selvan2020:TensorNetworksMedical}, audio \cite{Reyes2020:MultiScaleTensorNetwork}, and particle events \cite{Araz2021:QuantuminspiredEventReconstruction, Felser2021:QuantuminspiredMachineLearning}, highlighting their utility across a range of problems, and for various scientific disciplines.

A key advantage of using MPS over conventional deep-learning models for ML is their inherent interpretability \cite{Aizpurua2024:TensorNetworksExplainable}.
With MPS, the learned correlations are explicitly encoded in the model's tensors as conditional probabilities, allowing one to directly extract meaningful insights into the model's behavior and its representation of dependencies within a dataset.
For example, analyzing the entanglement properties of an MPS trained on MNIST image data allowed researchers to identify which features were most crucial for identifying samples \cite{Bai2022:UnsupervisedRecognitionInformative}.
In unsupervised anomaly detection, MPS achieved performance competitive with flexible deep-learning data modeling algorithms, such as variational autoencoders (VAEs) and generative adversarial networks (GANs), while offering considerably richer explainability \cite{Aizpurua2024:TensorNetworksExplainable}.
By computing interpretable properties of the MPS such as the von Neumann entropy and mutual information, researchers uncovered the informational content of individual features and their contextual dependencies, shedding light on which features were most important for identifying anomalous patterns.
This transparency provides a stark contrast to the opaque nature of deep-learning architectures, which, while highly expressive, are often challenging to understand and interpret \cite{Guidotti2018:SurveyMethodsExplaining}.

The quantification of complex correlations in time series is central to data-driven time-series modeling, which underpins a range of time-series ML tasks, from diagnosing disease from measured brain signals, to forecasting inflation rates from economic data.
There are many existing algorithmic approaches for tackling time-series analysis tasks, including time-series modeling, classification, anomaly detection, imputation, and forecasting \cite{Fulcher2018:FeaturebasedTimeseriesAnalysis}.
These methods range from traditional statistical methods, such as autoregressive (AR) models, which are tractable but can capture only relatively simple temporal structures, to highly flexible and powerful so-called `black-box' approaches based on deep neural networks, which can capture much more complex and long-range temporal structures (but are challenging to interpret).
The latter approach encompasses a range of powerful and flexible time-series ML algorithms, including transformers \cite{wenTransformersTimeSeries2023}, variational autoencoders (VAEs) \cite{girinDynamicalVariationalAutoencoders2021}, generative adversarial networks (GANs) \cite{zhang_comprehensive_2022}, and denoising diffusion probabilistic models \cite{yangSurveyDiffusionModels2024}.
Univariate time series, considered here to be uniformly sampled (at a constant sampling period $\Delta t$), and thus representable as the vector $x_t$ (for $t = 1, \dots, T$), encode potentially complex one-dimensional (temporal) correlation structures, similar to the one-dimensional (spatial) correlation structures of wavefunctions in one-dimensional quantum systems that can be approximated using MPS.
The success of MPS in approximating wavefunctions in one-dimensional quantum systems and recent high-performing ML applications motivates us to extend this promising and flexible algorithmic framework to tackle time-series ML problems.
In this paper, we develop and apply an MPS-based algorithm, MPSTime, for learning a joint probability distribution directly from an observed time-series dataset.
We show that the MPS ansatz is an efficient and powerful one for capturing a diverse range of complex temporal correlation structures, and further show how it can be used as the basis for novel algorithms that we introduce for tackling time-series ML problems, focusing here on time-series classification (the inference of a categorical label from a time series) and imputation (the inference of a subset of unobserved values from a time series).
We implement this algorithm and provide our classification and imputation tools in the publicly available software package \textit{MPSTime}\footnote{\url{https://github.com/hugopstackhouse/MPSTime.jl}}.

Extending the existing MPS-based ML framework of \citet{stoudenmireSupervisedLearningQuantumInspired2017} -- which was applied to two-dimensional images -- to univariate time series requires substantial further development.
While MNIST figures are characterized by grayscale pixels with approximately continuous grayscale values, significant progress can be made by approximating the pixels as either black or white \cite{Han2018:UnsupervisedGenerativeModeling}.
On the other hand, for time-series data, it is crucial to properly represent the real-valued amplitudes, $x_t$, in order to appropriately encode the statistical structure contained in the dataset.
This means that a carefully chosen encoding process is required to map each real-valued time-series datapoint into a finite dimensional vector that can be interfaced with MPS.
In fact, using an appropriate encoding, it was recently shown that the MPS ansatz can approximate any probability density function for continuous real-valued data \cite{Meiburg2023:GenerativeLearningContinuous}.
A second key aspect of our MPSTime algorithm is that we can use the same loss function for training the MPS both for classification and generative ML (e.g., synthetic time-series generation, forecasting, or imputation).
This means that MPSTime can be used to infer a probabilistic model of the joint probability distribution of a time-series data class, which can be used as the basis of a time-series classification model.
This unified framework for time-series ML stands in contrast to the existing algorithmic literature on time-series analysis that, due to the typical intractability of learning high-dimensional joint distributions from data, is highly disjoint in its development of different algorithms for different classes of statistical time-series problems.

This paper is structured as follows.
In Sec.~\ref{sec:mps-formalism} we cover the relevant MPS theory and specify the tensor network structure we used in subsequent investigations. 
In Sec.~\ref{sec:encoding-methodology} we present an approach for encoding real-valued time-series data into a finite-dimensional Hilbert space with nonlinear feature maps. 
Section~\ref{sec:imputation} and Sec.~\ref{sec:classification} describe the methods we use to perform time-series imputation and classification, respectively.
In Sec.~\ref{sec:experiments}, we firstly synthesize datasets to demonstrate the generative modeling and imputation capabilities of MPSTime. 
We then introduce three real-world datasets focusing on key application areas --energy, medicine, and physics -- to showcase the ability of MPSTime to perform both classification and imputation of missing data points.
Moving further, in Sec.~\ref{sec:interpretability} we highlight the interpretability of the joint probability distribution encoded in a trained MPS by sampling trajectories from the trained model and calculating conditional entanglement entropies.
By performing imputation on out-of-distribution instances, we also shed light on the representation of the data distribution learned by the MPS.
Broader implications of the work are discussed in Sec.~\ref{sec:discussionAndConclusion}.

\section{The Matrix-Product State Formalism}
\label{sec:mps-formalism}
In quantum many-body physics, a system of $T$ particles can be modeled by a wavefunction $\Psi(s_1, s_2,\ldots, s_T)$, which assigns a probability amplitude to each possible configuration of the many-body system. 
For any specific configuration $\mathbf{s} = (s_1, s_2, \ldots, s_T)$, Born's rule states that the squared-norm of the wavefunction i.e., $p(\mathbf{s}) = \lvert \Psi(\mathbf{s}) \rvert ^2$ defines the probability density of observing the configuration.
Here, the wavefunction encapsulates the full \textit{joint distribution} over an exponentially large space of $d^T$ possible states, where $d$ is the number of local states that each particle can take (e.g., $d = 2$ for spin-1/2 particles).

\begin{figure}
    \centering
    \includegraphics[width=1.0\linewidth]{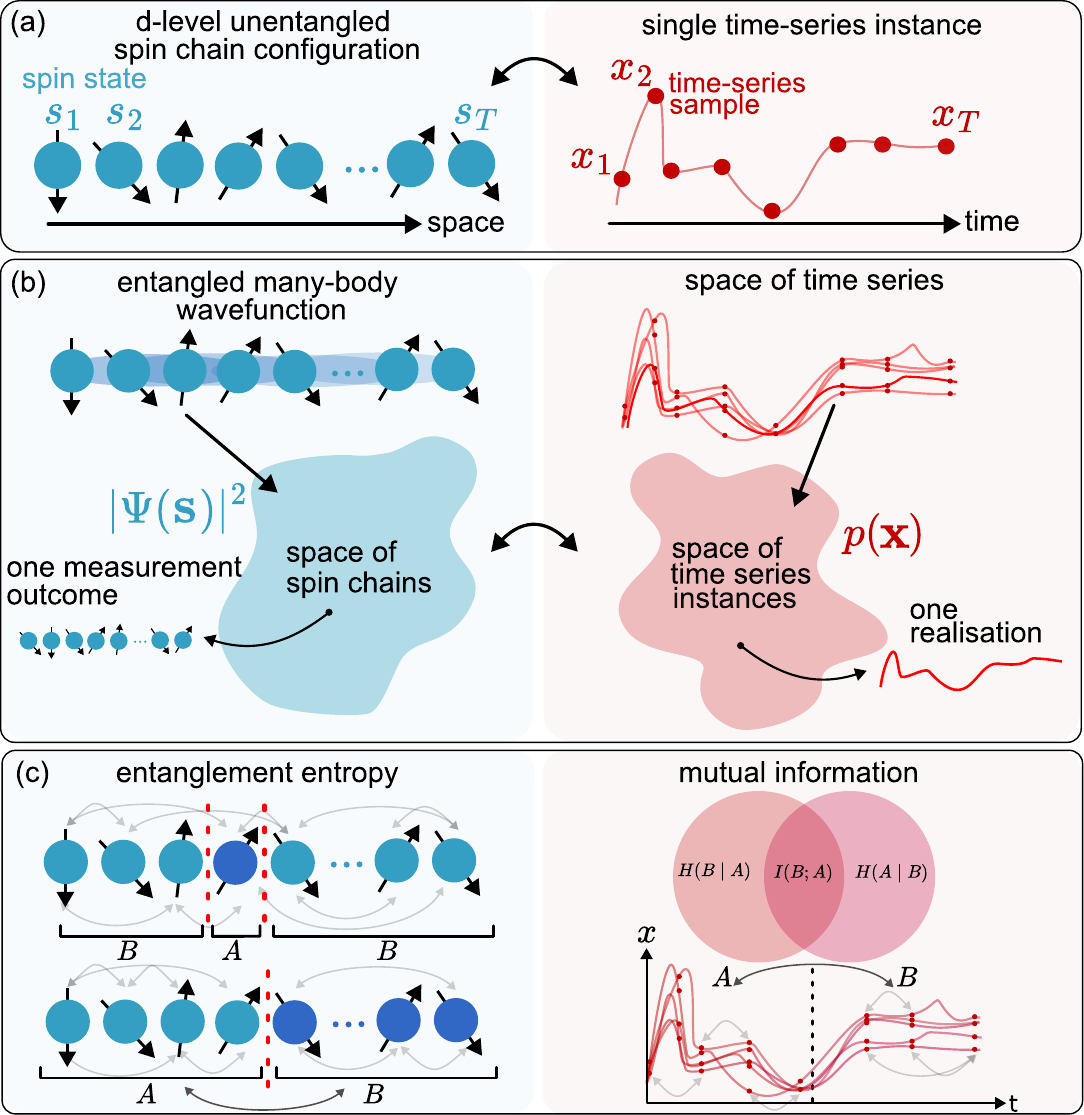}
    \caption{\textbf{Mapping between quantitative formulations of quantum many-body physics and time-series analysis.}
    Concepts in quantum physics (left, blue shading) and time-series analysis (right, red shading), and their associated theoretical and analytic tools, share similarities that motivate our MPS-based approach to time-series ML.
    \textbf{(a)} A $d$-level spin chain (left) exhibits one-dimensional (1D) spatial ordering, analogous to the 1D temporal ordering of time series (right). In this work, we propose mapping each time-series sample to a discrete state, similar to an individual $d$-level quantum spin. 
    \textbf{(b)} The probability density captured by the square of the wavefunction $|\Psi(\mathbf{s})|^2$ (left), is analogous to the joint probability density $p(\mathbf x)$ (right), which assigns a probability to every possible time series. 
    Here, the square of the wavefunction encodes a probability distribution over the space of possible measurement outcomes, while the joint density $p(x)$ defines a distribution over the space of possible time series (i.e., $\mathbb{R}^T$).
    A single measurement outcome (i.e., one spin chain configuration $\mathbf{s}$) governed by Born's rule corresponds to sampling a single time-series realization (i.e., one time-series instance $\mathbf{x}$) from its generative distribution.
    \textbf{(c)} The entanglement entropy (left), which quantifies the degree of quantum entanglement between two spatial subsystems (e.g., $A$ and $B$), is conceptually related to classical quantities such as the mutual information (right) between temporal segments (e.g., $A$ and $B$), both capturing statistical dependencies in their respective domains.
    }
    \label{fig:conceptual-picture}
\end{figure}

As we will show, for the purposes of time-series analysis, it is advantageous to view a $T$-sample time-series instance $\mathbf{x} = (x_1, x_2, \dots, x_T)$ as a specific configuration of a $T$-body system, as in Fig.~\ref{fig:conceptual-picture}(a), where each temporal observation $x_t$ -- originally real-valued and continuous -- is mapped to a discrete state $s_t$. 
Throughout the text, we refer to a single temporal observation $x_t$ as a time-series amplitude or time-series value.
Here, analogous to a quantum many-body system, the data wavefunction $\Psi(x_1, x_2,\dots,x_T)$ encodes the joint distribution of time-series data, and is determined by the properties of the underlying generative process. 
Analogous ideas in quantum many-body physics and time-series machine learning are illustrated in Fig.~\ref{fig:conceptual-picture}.
In this probabilistic picture, each time-series instance $\mathbf{x}$ of $T$ measurements thereby serves as a point sample from a $T$-dimensional joint distribution, as depicted in Fig.~\ref{fig:conceptual-picture}(b).
As with its quantum analogue, within our MPS formalism for ML the probability density of observing a time-series instance $\mathbf{x}$ from the distribution encoded by $\Psi$ is given by Born's rule:
\begin{equation}
\label{eq:born}
    p(\mathbf{x}) = \lvert \Psi(\mathbf{x}) \rvert ^2\,.
\end{equation}
Although such a wavefunction is exponentially large in general, it may be approximated by a $T$-site MPS ansatz \cite{schollwoeck_density-matrix_2011} which provides a compressed representation:
\begin{equation}
    \label{eq:MPS_formalism}
    W_{s_1, ..., s_T} = \sum_{\bm{\alpha}} A_{\alpha_1}^{s_1} A_{\alpha_1, \alpha_2}^{s_2} \dots A_{\alpha_{T-2}, \alpha_{T-1}}^{s_{T-1}} A_{\alpha_{T-1}}^{s_T}\,.
\end{equation} 
Here, $W$ is the low-rank MPS approximation of the original wavefunction (which can also be expressed in tensor form), each site $A^{s_t}$ corresponds to a measured point in time $x_t$, and the dimension of the bond indices $\bm{\alpha} = \{\alpha_1, \ldots, \alpha_{T-1}\}$ may be adjusted to tune the maximum complexity of the MPS ansatz.
Crucially, by compressing the wavefunction, and by extension, the joint distribution it encodes, the MPS approximation offers an explicit and tractable model of the generative distribution underlying observed time-series data.
Working directly with an approximation of the time-series joint distribution -- rather than relying solely on the empirical distribution of observations -- is what sets our MPS-based approach apart from existing methods.
In particular, we show how inferring a joint distribution of time series with MPS, analogous to the joint distribution of states encoded by the quantum many-body wavefunction, allows us to tackle important statistical learning problem classes, including classification and imputation.
Further, as in Fig.~\ref{fig:conceptual-picture}(c), by drawing on the breadth of existing tools for analyzing MPS in quantum many-body physics, we show that
interpretable quantities that would otherwise be highly challenging or impossible to obtain from the empirical dataset alone, can be straightforwardly extracted from the MPS.

\begin{figure*}
\includegraphics[width=\textwidth]{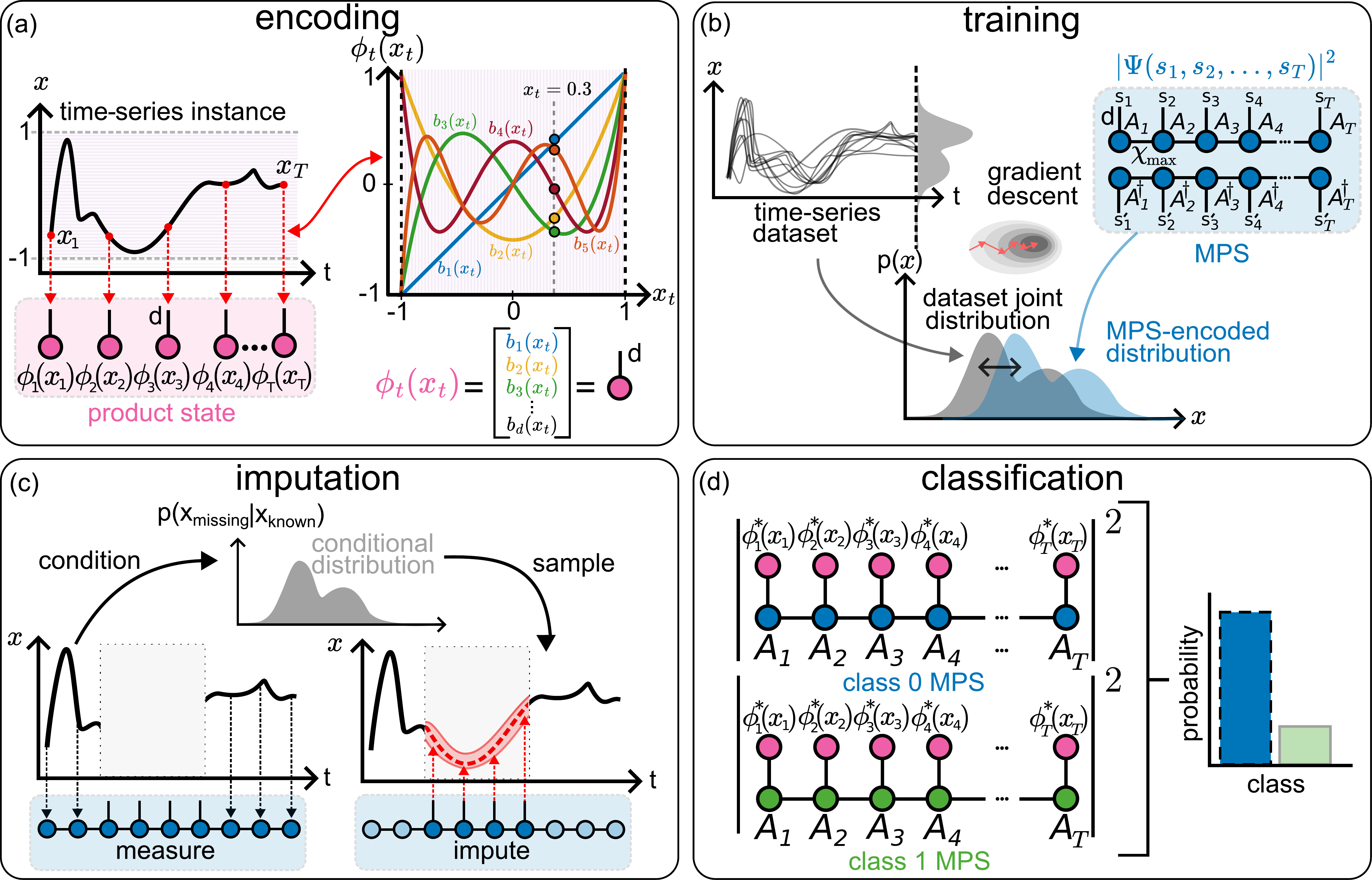}
\caption{\label{fig:wide}\textbf{MPSTime, a framework for time-series machine learning with Matrix-Product States (MPS).}
\textbf{(a)} Encoding: Each real-valued time series amplitude $x_t$ is encoded in a $d$-dimensional vector $\phi_t$ by projecting its value onto a truncated orthonormal basis with $d$ basis functions.
An entire time series (of length $T$ samples) is then encoded as a set of $T$ $\phi_t$ vectors, which we represent as a product state embedded in a $d^{T}$ dimensional Hilbert space. 
\textbf{(b)} MPS training: Using observed time series from a dataset, a generally entangled MPS -- depicted here using Penrose graphical notation -- with maximum bond dimension $\chi_{\rm{max}}$ is trained with a DMRG-inspired sweeping optimization algorithm to approximate the joint distribution of the training data.
Two copies of the trained MPS (one conjugate-transposed, denoted by the dagger $\dagger$) with open physical indices encodes the learned distribution, allowing us to sample from and do inference with complex high-dimensional time-series distributions.
In this work, we introduce MPS-based learning algorithms, which we collectively refer to as MPSTime, for two important time-series ML problems: \textbf{(c)} 
imputation (inferring unmeasured values of a time series), and \textbf{(d)} classification (inferring a time-series class).
\textbf{(c)} Generative time-series modeling: we use conditional sampling to perform imputation of missing datapoints.
Known points of a time series (black lines) project the MPS into a subspace, which is then used to find the unknown datapoints (red line).
The same method can be used to tackle some forecasting problems if the missing points are future values.
\textbf{(d)} MPS for classification: multiple labeled classes of time series are used to train MPSs.
Taking the overlap of unlabeled time-series data (encoded as a product state) with each MPS determines its class.
}
\end{figure*}

\subsection{Encoding time series as product states}
\label{sec:encoding-methodology}
The MPS ansatz is a powerful tool for compressing functions in an exponentially large space. 
To represent the joint probability distribution underlying a time-series dataset using MPS, the continuous-valued time-series amplitudes, $x_t$, must first be mapped to vectors in a finite-dimensional Hilbert space.
To achieve this mapping, we employ $d$-dimensional nonlinear feature maps, which throughout the text we also refer to as the `encoding': 
\begin{equation}
    \label{eq:basis-fns}
    \phi_t(x_t) = [b_{1}(x_t), b_{2}(x_t),\ldots, b_{d}(x_t) ]_t\,,
\end{equation} 
which for a given choice of real or complex basis functions $b_i$, map a real valued time-series amplitude $x_t$ to a vector in $\mathbb{R}^d$ or $\mathbb{C}^d$, respectively. 
This mapping via the encoding $\phi_t(x_t)$ is depicted schematically in Fig.~\ref{fig:wide}(a).
In general, the choice of basis functions $[b_1, \ldots, b_d]_t$ can be time dependent, but here we consider a time-independent feature map.
When applied to an entire time-series instance of $T$ samples, the complete feature map $\Phi(\mathbf{x})$ is then given by the tensor product of local feature maps:
\begin{equation}
    \Phi(\mathbf{x}) = \phi_1(x_1) \otimes \phi_2(x_2) \otimes \ldots \otimes \phi_T(x_T)\,.
\end{equation}

To ensure the probabilistic interpretation of the wavefunction parameterized by the the Matrix-Product State (MPS) $W$ in Eq.~\eqref{eq:MPS_formalism} remains valid, the normalization condition:
\begin{equation}
    \label{eq:norm_condition}
    \int_{x} |\Phi(\mathbf x) \cdot W|^2 \ud  \mu(x) = 1\,,
\end{equation}
must be satisfied \cite{stoudenmireSupervisedLearningQuantumInspired2017}.
This condition will hold, provided that: (i) the MPS is unitary; and (ii) the feature map is orthonormal under the measure $\mu(x)$
\cite{Meiburg2023:GenerativeLearningContinuous}.
While many valid choices of feature map exist, including the Fourier basis, Laguerre and Hermite polynomials, in the present work we choose $b_i(x) = P_i(x)$, the $i$-th orthonormal Legendre polynomial.
The Legendre polynomials are well suited to encoding time-series data, as they satisfy the orthonormality conditions on the compact interval $[-1, 1]$, and can generalize to any physical dimension $d$.
Higher physical dimension $d$ gives rise to larger tensors which, coupled with larger bond dimension, allows the MPS to express more complex inter-site interactions. 
Such increased model complexity, however, comes at the cost of additional computational storage and runtime. 

To encode real-valued time-series amplitudes using the Legendre basis, we first apply a pre-processing step to map the original values in the data domain $\mathbf{x} \in \mathbb{R}$ to a compact encoding domain $ \mathbf{x} \in [-1,1]$.
When using the MPS for time-series classification in Sec.~\ref{section:classification-results} we apply a scaled outlier-robust sigmoid transformation \cite{Fulcher2013:HighlyComparativeTimeseriesa}, before transforming the dataset to map the dataset minimum to $-1$, and the dataset maximum to 1 (via min--max normalization).
For imputation, preserving the shape of the data is essential, and thus care must be taken to ensure minimal distortion to imputed time series.
For this reason, we avoid nonlinear transformations (e.g., sigmoid transform), which can amplify encoding-related errors, particularly when mapping from the encoding domain to the original data domain i.e., $[-1, 1] \rightarrow \mathbb{R}$.
Instead, for time-series imputation, we apply a linear min--max normalization as the only pre-processing step.
Further details about the specific data transformations we use are in Appendix~\ref{appendix:data-pre-proc-details}.



\subsection{Training an MPS-based generative model} \label{sec:MPStraining}
In the previous section, we described how continuous-valued time-series instances can be converted into a discrete representation suitable for MPS-based machine learning. 
We now detail the procedure used to train an MPS in order to model the joint distribution defined by a finite set of time-series instances sampled from it. 
As we will show, the resulting MPS can then be used to solve diverse time-series machine learning problems, such as imputation of missing data, classification of unseen data, and generation of new data, within a general framework.

\subsubsection{Loss function for generative modeling}
In order to approximate the joint distribution $p(\mathbf{x})$ of the data, an MPS model is trained on a finite number of encoded time-series instances sampled from it, shown schematically in Fig.~\ref{fig:wide}(b).
Starting with a randomly initialized MPS, the model is trained by iteratively updating its tensor entries to minimize the Kullback--Leibler (KL) divergence \cite{Kullback1951:InformationSufficiency, Meiburg2023:GenerativeLearningContinuous, Han2018:UnsupervisedGenerativeModeling, Li2018:ShortcutMatrixProduct, Sun2020:TangentSpaceGradientOptimization}, a measure of the discrepancy between the probability distribution $p(\mathbf{x})$ captured by the MPS, and the true distribution $q(\mathbf{x})$ from which the training data are sampled:
\begin{equation}
\label{eqn:DKL}
    D_{KL} = \int_\mathbf{x} q(\mathbf{x}) \log{\frac{q(\mathbf{x})}{p(\mathbf{x})}} \ud \mathbf{x}\,,
\end{equation}
where the integral is over every possible configuration of $\mathbf{x}$. 
In practice, the true distribution $q$ is unknown, so it is common to minimize an averaged negative log-likelihood (NLL) loss function over the set of available training time-series instances $N$ \cite{Meiburg2023:GenerativeLearningContinuous, Liu2023:TensorNetworksUnsupervised, Mossi2024:MatrixProductState, Cheng2019:TreeTensorNetworks},
\begin{equation} \label{eq:NLL}
    \mathcal{L}_G = -\frac{1}{N}\sum_{n=1}^{N} \log{ p(\mathbf{x}_n) }\,.
\end{equation}
This is equivalent to minimizing the KL divergence (up to a constant). 
Once the MPS is trained, the joint probability density $p(x)$ is extracted from the MPS using the Born rule, which now takes the form,
\begin{equation}
p(\mathbf{x}_n) = \lvert W \cdot \Phi(\mathbf{x}_n) \rvert^2\,.
\end{equation}
To extend training to classification, we treat each class of time series as originating from a separate joint probability distribution. 
In other words, given $L$ classes of data, classification requires learning $L$ probability distributions with $L$ unique MPS models. 
As in~\citet{stoudenmireSupervisedLearningQuantumInspired2017}, an equivalent, yet more efficient approach is to attach an $L$-dimensional \textit{label index} to a single site of an MPS:
\begin{equation}
\label{eq:MPS_formalism_label}
    W^l_{s_1, ..., s_T} = \sum_{\bm{\alpha}} A_{\alpha_1}^{s_1} A_{\alpha_1, \alpha_2}^{s_2} \ldots A^{l, s_j}_{\alpha_{j-1}, \alpha_{j}}\ldots A_{\alpha_{T-1}}^{s_T}\, ,
\end{equation}%
and minimize the sum of the losses across every class:
\begin{equation}
\mathcal{L} = - \frac{1}{N} \sum_{n=1}^{N} \sum_{l=1}^{L}  \log{\lvert W^l \cdot \Phi(\mathbf{x}_n) \rvert ^2}\,.
    \label{eq:NLL-multi}
\end{equation}
The addition of a label index $l$ allows us to approach generative modeling and classification in a unified manner, as the total loss $\mathcal{L}$ reduces to Eq.~\eqref{eq:NLL} in the single class `unsupervised' case. 
To avoid confusion about class dependence in the generative modeling sections, we will omit the label index $l$ from our notation when it is one dimensional (i.e., single class).






\subsubsection{Sweeping optimization algorithm}
Using the loss function $\mathcal{L}$ in Eq.~\eqref{eq:NLL-multi} to quantify the MPS approximation of the data joint distribution, we now outline the local optimization algorithm we used to minimize its value on the training data. 
We use the same DMRG-inspired sweeping algorithm in~\citet{stoudenmireSupervisedLearningQuantumInspired2017}, but with a modified version of tangent-space gradient optimization (TSGO) \cite{Sun2020:TangentSpaceGradientOptimization} for the local tensor update step.
Here, we present only the key algorithmic steps as a summary, and direct readers to \citet{stoudenmireSupervisedLearningQuantumInspired2017} for a detailed treatment.

Starting with an MPS of $T$ sites, with each site being represented by a tensor of random entries, fixed encoding dimension $d$, and uniform initial bond dimension $\chi_{\rm init}$.
\begin{enumerate}[noitemsep]
    \item Attach the label index $l$ to the rightmost site of the MPS, and place the MPS in left-canonical form \cite{schollwoeck_density-matrix_2011}.
    \item Merge the rightmost pair of tensors in the MPS, $A^{l, s_{T}}_{\alpha_{T-1}}$ and $A^{s_{T-1}}_{\alpha_{T-1} \alpha_{T-2}}$, to form the \textit{bond tensor} $\mathcal{B}$.
    \item Holding the remaining $T-2$ tensors fixed, update $\mathcal{B}$ using a modified TSGO steepest descent rule: 
 
    \begin{equation}
        \mathcal{B}' = \mathcal{B} - \eta \frac{
            \partial \mathcal{L}/\partial \mathcal{B}
        }{
        \left| \partial \mathcal{L}/\partial \mathcal{B} \right|
        }\,,
    \end{equation} 
    where $\eta$ is the \textit{learning rate}. Since $\mathcal{L}$ is not complex differentiable, we use (one-half times) the Wirtinger derivative \cite{kreutz2009complex} to compute $\frac{\partial \mathcal{L}}{\partial \mathcal{B}}$, thereby accommodating complex-valued encodings (e.g., Fourier basis) if required.
    \item Normalize the updated bond tensor: 
    \begin{equation}
        \mathcal{B}' \leftarrow \frac{\mathcal{B'}}{\lvert{\mathcal{B'}\rvert}}\,.
    \end{equation}
    \item Decompose $\mathcal{B}'$ back into two tensors by singular value decomposition (SVD), retaining at most $\chi_{\rm{max}}$ singular values.
    The number of retained singular values corresponds to the maximum bond dimension shared between the two updated sites. 
    The left and right tensors of the SVD are chosen so that the label index moves along the MPS, and is part of the bond tensor $\mathcal{B}$ at every step.
    \item Repeat steps 2-5 for each adjacent pair of remaining tensors, traversing from right-to-left, then from left-to-right, for a fixed number of iterations, or until convergence of the loss function $\mathcal{L}$ is achieved.
\end{enumerate}
As in \citet{stoudenmireSupervisedLearningQuantumInspired2017}, we define one \textit{sweep} of the MPS to be a `round-trip', i.e., one backward and one forward optimization pass through all tensors in the MPS.

\subsection{Time-series imputation algorithm} \label{sec:imputation}

\begin{figure*}
    \centering
    \includegraphics[width=0.95\linewidth]{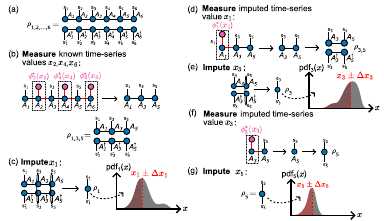}
    \caption{\textbf{An MPS-based algorithm for time-series imputation.}
    Here we consider an illustrative example of an imputation problem involving a six-site MPS, represented graphically with Penrose notation \cite{penrose1971applications}, where the time-series values $x_2, x_4, x_6$ are observed, and we would like to impute the unobserved values $x_1, x_3, x_5$.
    \textbf{(a)} Two copies of the trained MPS (one conjugate-transposed, indicated by the dagger, $\dagger$) with open physical indices encode the joint distribution over all possible states, given by $\rho_{1,2,\dots,6}$ (Eq.~\eqref{eq:outer-joint}).
    \textbf{(b)} The MPS is projected into a subspace where the states $s_2$, $s_4$, $s_6$, corresponding to each of the known time-series values, have been measured.
    The updated MPS now encodes the joint distribution over the remaining states $s_1$, $s_3$, $s_5$, conditional upon having measured $s_2, s_4, s_6$.
    \textbf{(c)} The single-site conditional reduced density matrix $\rho$ (as in Eq.~\eqref{eq:ssrdm}) is obtained by tracing over all remaining unmeasured sites.
    By evaluating the probability density function $\textrm{pdf}_i(x) = \phi_i^\dagger(x) \rho_i \phi_i(x)$ for $x$ in the encoding domain, we estimate the first unobserved value $x_1$ and its uncertainty $\Delta x_1$ using the \textit{median} (see Eq.~\eqref{eq:pdf_median}) and weighted median absolute deviation (WMAD) of the probability density function, respectively.
    \textbf{(d)} The MPS is projected onto the estimated state $s_1 = \phi_1(x_1)$ (Eq.\eqref{eq:ssrdm}), yielding an updated MPS which encodes the joint distribution over remaining unmeasured states $s_3, s_5$.
    \textbf{(e, f, g)} We repeat the process of estimating the next missing value (i.e., $x_3$) using the median of the corresponding pdf (conditioned on all previously known and imputed values), then projecting the MPS onto the estimated state (i.e., $s_3 = \phi_3(x_3)$), until all remaining unobserved values are recovered.
    }
    \label{fig:imputation-summary}
\end{figure*}

Many empirical time series contain segments of time for which values are missing (e.g., due to sensor drop-out) or contaminated (e.g., with artifacts).
This complicates analyses and, for many algorithms (such as classification or dimension-reduction), requires \textit{imputation}, i.e., that the missing data be inferred (i.e., `filled in') from values that were observed \cite{Wang2024:DeepLearningMultivariate, Fang2020:TimeSeriesData}.
For example, in some astrophysical star surveys, periods of sensor downtime is unavoidable due to operational procedures \cite{Garcia2014:ImpactAsteroseismicAnalyses}, leaving `gaps' in the recorded light curves of stars during which no observations are made.
Existing approaches to imputing missing values in time series range from filling them with a simple statistic of observed values (e.g., the mean of the full time series \cite{Osman2018:SurveyDataImputation}), linearly interpolating across the missing data, to inferring missing values based on fitted linear models (e.g., autoregressive integrated moving average (ARIMA)), similarity-based methods (e.g., $k$-nearest neighbor imputation from a training set of data (K-NNI) \cite{Beretta2016:NearestNeighborImputation}), and machine learning algorithms (e.g., generative adversarial networks (GANs) \cite{Luo2018:MultivariateTimeSeries}, variational auto-encoders (VAEs) \cite{McCoy2018:VariationalAutoencodersMissing}).
In this work, we present a generative approach to time-series imputation by leveraging the data distribution modeled by the MPS $W$ (as in Eq.~\ref{eq:MPS_formalism}), summarized schematically in 
Fig.~\ref{fig:imputation-summary} using Penrose graphical notation \cite{penrose1971applications}.
Our imputation approach, shown schematically in Fig.~\ref{fig:wide}(c), consists of two main steps: (i) conditioning the MPS on observed values, and (ii) using the resulting conditioned MPS to infer missing values.
Here, we summarize the key algorithmic steps and refer the reader to Appendix~\ref{app:interp_alg} for full details.

Let the ordered sequence $\mathbf{x}^{*} = x_1, x_2, \ldots, x_T$ represent a time series with observed values $\{\mathbf{x}^o: o \in O\}$, and missing values $\{\mathbf{x}^m: m \in M\}$, where $M := \{1,2,\ldots,T\} \setminus O$.
The reduced density matrix for the unknown states $\rho_M$ can be obtained by first projecting the observed states onto the MPS:
\begin{equation} \label{eq:MPScondition}
    \widetilde{W}_{s_{m_1}, s_{m_2}, \ldots } =   W_{s_1, ..., s_T} \cdot \prod_{i \in O} \frac{\phi_i^{s_i \dagger}(x_i)}{\sqrt{P(x_i)}}\,,
\end{equation}
and then taking the outer product:
\begin{equation} \label{eq:outer-joint}
    \rho_M = \widetilde{W} \widetilde{W}^\dagger \,,
\end{equation}
where $\widetilde{W}^\dagger$ is the conjugate transpose of $\widetilde{W}$.
Throughout the text, we refer to the process of projecting the MPS onto an observed value as either a measurement or conditioning a joint distribution on an observed value. 
The scale factor $\sqrt{P(x_i)}$ ensure that the MPS remains correctly normalized throughout.
Technical details of the implementation are described fully in Appendix~\ref{app:interp_alg}.

An important feature of MPS-based imputation is that each missing value is imputed one at a time, using both observed values and imputed values that were previously missing. 
Therefore, the outcome of an imputation depends on the order in which it is performed. 
As time series are indexed in time order, we choose to perform imputation sequentially from earliest to latest using the `chain rule' of probability, as described in \cite{Ferris2012:PerfectSamplingUnitary}.  
While previous approaches typically sample a random state from the single-site conditional distributions (e.g., via inverse transform sampling \cite{Meiburg2023:GenerativeLearningContinuous}), we use a deterministic approach wherein we select the state that corresponds to the \textit{median} of the conditional distribution at each imputation site, as in Fig.~\ref{fig:imputation-summary}(c). 
Our choice to use the median is motivated by the observation that discretizing continuous time-series values using a finite number of basis functions $d$ induces an artificial broadening of the MPS-approximated probability distribution.
Sampling directly from this broadened distribution can produce values that are unlikely under the true data distribution, leading to degraded imputation performance.
As detailed in Appendix~\ref{appendix:enc-error}, selecting the median of the distribution effectively mitigates this broadening effect and provides a robust best estimate for each imputed value. 
In what follows, we present our MPS-based algorithm for time-series imputation, shown schematically in Fig.~\ref{fig:imputation-summary}.
Starting at the earliest missing site, $i$, in $M$:
\begin{enumerate}[noitemsep]
    \item Obtain the single-site reduced density matrix (RDM) for site $i$, $\rho_i$, by taking the partial trace over all remaining (unmeasured) sites except for site $i$:
    \begin{equation} \label{eq:ssrdm}
        \rho_i = \operatorname{Tr}_{M \setminus \{i\}} \left( \rho_M \right) = \operatorname{Tr}_{M \setminus \{i\}} \left( \widetilde{W} \widetilde{W}^\dagger \right)\,,
    \end{equation}
    where $\widetilde{W}$ is the projected MPS incorporating the observed states and $\rho_M$ is its corresponding density matrix as in Eq.~\eqref{eq:outer-joint}.
    \item Using the RDM, compute the cumulative distribution function $F_i(x)$ as 
    \begin{equation}
            F_i(x) = \frac{1}{Z} \int_{-1}^{x} \phi_i^\dagger(x') \rho_i \phi_i(x') \ud x'\,,
    \end{equation}
    with normalization factor $Z$ chosen so that $F_{i}(1) = 1$.
    \item Infer the unobserved value $x^*_i$ using the median of the conditional distribution as:
    \begin{equation} \label{eq:pdf_median}
            x^*_i = \operatorname*{arg\,min}_x \left|F_i(x) - \frac{1}{2} \right|\,.
    \end{equation}
    \item Project the MPS onto the selected state $s_i$ corresponding to the inferred value $x^*_i$ by contracting the MPS with $s_i = \phi_i^\dagger(x_i^*)$, as in Eq.~\eqref{eq:MPScondition}:
    \begin{equation} \label{eq:site-projection}
        \widetilde{W}' =  \widetilde{W} \cdot \frac{\phi_i^\dagger(x^*_i)}{\sqrt{P(x^*_i)}}\,.
    \end{equation}
    \item Proceed to the next unobserved value's MPS site and repeat steps 1--4 using the lower-rank MPS $\widetilde{W}'$ until all unobserved values $\mathbf{x}^m$ are recovered.
\end{enumerate}

\subsection{Time-series classification algorithm} \label{sec:classification}
Building on the formalism in Sec.~\ref{sec:MPStraining}, we now describe how the MPSTime algorithm can be extended to the problem of time-series classification.
Specifically, for a dataset of time-series instances, each associated with a distinct class label $l$ from a finite set of classes $L$, we train a model that can accurately predict the class label $l \in L$ of new, unseen time-series instances using the learned generative distributions of the classes.

As described in Sec.~\ref{sec:MPStraining}, classification is achieved by training $L$ distinct MPSs.
For each class indexed by $l$, we train a separate MPS $W$ on the subset of instances belonging to $l$.
To classify an unseen time-series instance $\mathbf{x}$, we then compute its (non-normalized) probability density $p(\mathbf{x})$ under each of the $L$ MPSs, and select the label $l$ of the MPS that assigns the highest relative probability to the instance, as illustrated schematically in Fig.~\ref{fig:wide}(d).

In practice, the MPS formalism allows for a more efficient solution: 
by attaching an $L$-dimensional label index to a single MPS (at any site), we can train a unified model $W^l$ to simultaneously encode the generative distributions for all classes.
For any time series $\mathbf{x}$, encoded as a product state $\Phi(\mathbf{x})$, we can then define the contraction:
\begin{equation}
    f^l(\mathbf{x}) = W^l \cdot \Phi(\mathbf{x})\,,
\end{equation}
where the model output $f^l(\mathbf{x})$ is a length $L$ vector, with each entry being the similarity (overlap) between $\mathbf{x}$ and class $l$. 
To obtain the probability of $\mathbf{x}$ belonging to each class, we take the squared-norm of the output vector in order to satisfy the Born rule:
\begin{equation}
    p(\mathbf{x} \mid l) = |f^l(\mathbf{x})|^2\,.
\end{equation}
As all classes are assumed to have the same prior probability $p(l)$, then the probability of each class given an unlabeled time series, $p(l\mid \mathbf{x})$, is directly proportional to its non-normalized likelihood $p(\mathbf{x} \mid l)$ i.e., $p(l \mid \mathbf{x}) \propto p(\mathbf{x} \mid l)$ (cf. Bayes' rule \cite{Bishop2006:PatternRecognitionMachine}).
Therefore, the predicted class can be determined directly from the model output by selecting the index $l$ corresponding to the largest entry in $f^l(\mathbf{x})$:
\begin{equation}
    \operatorname*{arg\,max}_l |f^{l}(\mathbf{x})|^2\,.
\end{equation}
%

In this section, we have introduced MPSTime, an MPS-based algorithm for estimating the joint distribution $p(\mathbf{x})$ underlying a dataset of time series.
By leveraging the properties of the MPS to encode an approximation of complex joint distributions, it is possible to sample from and do inference on those distributions, allowing it to form the basis for novel time-series machine-learning problems including imputation and classification.
Crucially, while existing algorithms often address different problem classes like classification and imputation using specifically tailored algorithms, MPSTime can tackle them within a unified time-series modeling framework.

\section{Numerical Experiments}
\label{sec:experiments}

To validate the proposed MPSTime framework developed above, in this section we conduct a range of numerical experiments aimed to demonstrate its ability to approximate time-series joint distributions, and apply this structure to practical time-series ML tasks.  
We begin in Sec.~\ref{sec:time-series-datasets} by detailing the construction of the datasets comprising the tasks.
Then, in Sec.~\ref{sec:imputation-experiments}, we empirically validate our framework for MPS-based imputation on these datasets, showing that MPSTime can successfully learn the underlying joint distribution $p(\mathbf{x})$ directly from time-series data.

\subsection{Time-series datasets} \label{sec:time-series-datasets}
To validate our MPS-based algorithm for time-series machine learning (ML), we investigated several synthetic and real-world datasets.
In this work, we focus on the tasks of time-series imputation and classification, selected to encompass the breadth of application of MPS-based inference for time series. 
Here, we detail the process to derive the datasets from publicly available repositories.
Table~\ref{tab:dataset-summary} provides a summary of the datasets used in our experiments.

\subsubsection{Synthetic datasets}
\label{sec:dataset-details-synthetic}

We first aimed to validate the ability of MPSTime to infer distributions over time-series data by simulating time-series instances from an analytic generative model.
To achieve this, we generated time series of length $T = 100$ samples from a `noisy trendy sinusoid' (NTS) model:
\begin{equation} \label{eq:noisy-trendy-sinusoid}
    x_t = \sin{\left(\frac{2\pi}{\tau}t + \psi\right)} + \frac{mt}{T} + \sigma n_t\,,
\end{equation}
where $x_t$ is the observed value at time $t \in [0, T]$, $\tau$ is the period of the sinusoid, $m$ is the slope of a constant linear trend, $\psi \in [0, 2\pi)$ is the phase offset, $\sigma$ is the noise scale, and $n_t \sim \mathcal{N}(0,1)$ are i.i.d. normally distributed random variables.
Time-series datasets were then constructed by repeatedly generating time-series instances as realizations from Eq.~\eqref{eq:noisy-trendy-sinusoid}.
The diversity of the time-series dataset (and thus complexity of the joint distribution over time series) can be controlled via the parameters of Eq.~\eqref{eq:noisy-trendy-sinusoid} that are fixed for all time series in the dataset, versus the parameters that are allowed to vary across the dataset.
We constructed five NTS datasets of differing complexity, generated through different numbers and degrees of parametric freedoms, as detailed below:
\begin{enumerate}[label=\roman*), nosep]
    \item \textbf{Simple setting:} For the simplest dataset, \textbf{NTS1}, we fixed the period ($\tau = 20$), trend ($m = 3$) and noise scale ($\sigma = 0.1$) in Eq.~\eqref{eq:noisy-trendy-sinusoid}, while allowing the phase $\psi$ to vary across the time-series dataset.
    This variation was achieved by sampling a value for $\psi$ for each time series as a random sample from a uniform distribution, as $\psi \sim U(0, 2\pi)$.
    We generated a training set containing $300$ time series, and a test set of $200$ time series from the same generative model.
    \item \textbf{Challenging settings:} To simulate time series exhibiting richer dynamical variation, and therefore more complex data distributions, we systematically expanded the parameter space of the NTS model.
    Fixing the noise scale at $\sigma = 0.1$, we constructed four datasets with progressively more complex data distributions by varying the trend $m$ and period $\tau$ parameters:
    (i) \textbf{NTS2}, with a single trend ($m = 3.0$) and three periods ($\tau \in \{20, 30, 40\}$);
    (ii) \textbf{NTS3}, with three trends ($m \in \{-3, 0, 3 \}$) and a single period ($\tau = 20$);
    (iii) \textbf{NTS4} with two trends ($m \in \{-3, 3\}$) and two periods ($\tau \in \{20, 40\}$);
    and (iv) \textbf{NTS5}, with three trends ($m \in \{-3, 0, 3 \}$) and three periods ($\tau \in \{20, 30, 40 \}$).
    For all datasets, the phase offset $\psi$ was drawn randomly from a uniform distribution $\psi \sim U(0, 2\pi)$, and the free parameter(s) were selected randomly from the set of allowed values.
    For NTS2, NTS3, and NTS4, we generated a training dataset containing $400$ time series (and $500$ for the most complex NTS5 model) to densely sample the larger model parameter space, and all models were evaluated on a test set of $200$ time series (generated from the same model and parameter settings).
    \end{enumerate}
\subsubsection{Real-world datasets} \label{sec:real-world-dataset-details}
To evaluate the performance of MPSTime on data for which the underlying generative process is unknown, we selected three real-world time-series datasets to analyze, representative of three key application domains: (i) medicine, (ii) energy, and (iii) astronomy:
\begin{enumerate}[label=\roman*)]
    \item \textbf{ECG:} This medical dataset, labeled as \texttt{ECG200} in the UCR TSC archive \cite{Dau2019:UCRTimeSeries}, is a two-class dataset of time series representing electrocardiogram (ECG) traces of single heart beats, labeled according to whether the signal originated from a healthy or diseased (myocardial infarction) patient \cite{Olszewski2001GeneralizedFE}.
    The dataset comprises 100 training instances and 100 test instances, where each time series contains $T = 96$ samples.
    Of the $200$ total time series, $133$ ($66.5\%$) are labeled as `healthy', and the remaining 67 ($33.5\%$) are labeled as `abnormal'.
    For classification and imputation tasks, we used the same open dataset published in the UCR time-series classification archive \cite{Dau2019:UCRTimeSeries}. 
    \item \textbf{Power Demand:} The Power Demand dataset, labeled as \texttt{ItalyPowerDemand} in the UCR repository \cite{Dau2019:UCRTimeSeries}, is a two-class dataset of time series corresponding to one day of electrical power demand in a small Italian city, sampled at 1\,h intervals \cite{Keogh2006:IntelligentIconsIntegrating}.
    Each time series is assigned to either `winter' (recorded between the months of October and March) or `summer' (recorded between the months of April and September).
    The dataset is split into $67$ training instances and $1029$ test instances, each of length $T = 24$ samples.
    Of the total $1096$ time series, 547 (49.9\%) are labeled as `winter' and the remaining 549 (50.1\%) are labeled as `summer'. 
    For classification and imputation, we use the same open dataset published in the UCR time-series classification archive \cite{Dau2019:UCRTimeSeries}.
    \item \textbf{Astronomy:} The Astronomy dataset is derived from the UCR TSC archive \texttt{KeplerLightCurves} dataset \cite{Dau2019:UCRTimeSeries} and comprises $1319$ light curves from NASA's Kepler mission \cite{Barbara2022:ClassifyingKeplerLight}.
    Individual time series correspond to stellar brightness measurements from a single star, sampled every $30$ minutes over a three-month quarter (`Quarter 9'), yielding length $T = 4767$ sample instances.
    For our classification task, we selected two of the seven classes that were particularly challenging to distinguish by eye which were (i) non-variables, and (ii) $\delta$ Scuti stars.
    We then applied random under-sampling to the majority class ($\delta$ Scuti stars, originally containing 411 instances) to yield a balanced class distribution (201 non-variable stars and 201 $\delta$ Scuti stars).
    Instances were randomly assigned to either the train or test set using an 80/20 train/test ratio, and each time-series instance was truncated to its first 100 samples for classification.

    For the imputation task, we selected two different classes from the \texttt{KeplerLightCurves} dataset: (i) RR Lyrae stars; and (ii) $\gamma$ Doradus stars.
    Unlike the Power Demand and ECG datasets, here we focused on the typical setting in which imputation is performed within a single time-series instance by using observed values to infer missing values.
    To construct the imputation datasets, we randomly selected five time-series instances from each class.
    These instances, each of length $T = 4767$ samples was truncated to $4700$ samples and segmented into 47 non-overlapping windows of $T = 100$ samples.
    Windows containing true missing data gaps were discarded, and 80\% of the remaining `clean' windows (43 windows) were randomly allocated to the training set, with 20\% (2 windows) allocated to the test set.
\end{enumerate}

\begin{table}[h!] 
\centering 
\caption{
\textbf{Summary of all simulated and real-world time-series datasets analyzed.}
Datasets are categorized into those used for classification (labeled `C') and imputation (labeled `I') tasks, see Sec.~\ref{sec:time-series-datasets} for details.
}
\label{tab:dataset-summary}
\begin{tabular}{ c c c c c }
    \hline
    \textbf{Dataset} & \textbf{Train Size} & \textbf{Test Size} & \textbf{Length}\\
    \hline
    ECG (C) & 100 (50\%) & 100 (50\%) & 96\\
    Power Demand (C) & 67 (6\%) & 1029 (94\%) & 24\\
    Astronomy (C) & 321 (80\%) & 81 (20\%) & 100\\
    NTS1 (I) & 300 (60\%) & 200 (40\%) & 100\\
    NTS2--4 (I)\textsuperscript{\S} & 400 (67\%) & 200 (33\%) & 100\\
    NTS5 (I) & 500 (71\%) & 200 (29\%) & 100\\
    ECG (I) & 100 (50\%) & 100 (50\%) & 96\\
    Power Demand (I) & 67 (6\%) & 1029 (94\%) & 24\\
    Astronomy (I)\textsuperscript{\textdaggerdbl} & 43 (96\%) & 2 (4\%) & 100\\ 
    \hline
\end{tabular}
\textsuperscript{\S}\footnotesize NTS2--4 refers to datasets NTS2, NTS3, and NTS4.\\
\textsuperscript{\textdaggerdbl}\footnotesize For consistency, we use `dataset' to refer to the collection of fixed-length windows extracted from a single time-series instance in the Astronomy (I) dataset.
A total of 10 such datasets (i.e., 10 independent and windowed time-series instances) were evaluated.
\end{table}

\subsection{Time-series imputation}
\label{sec:imputation-experiments}
The recovery of missing time-series values via imputation represents a key challenge in time-series analysis, as real-world data collection rarely occurs under ideal conditions.
Sensor dropouts due to malfunction, operational constraints, or other disruptions, can give rise to contiguous blocks of unobserved time-series values.
Robust imputation methods that accurately reconstruct missing values, while preserving the underlying temporal structure of the data, are essential for reliable analysis and inference on such datasets.

In this work, we specifically focus on univariate time-series imputation in a dataset-level setting, where a model is trained on a collection of $N$ `clean' (non-corrupted) time-series instances of fixed length $T$.
At test time, imputation is performed on unseen, partially corrupted instances of the same length $T$ by conditioning on the observed values, and reconstructing unobserved values from the joint distribution learned on the training set.
We note that this setup differs from the typical univariate imputation setting, where reconstruction is performed within a single, arbitrarily long time series, and using only its own observed values \cite{moritzComparisonDifferentMethods2015, sanwouo_ts-pothole_2024}.

We now turn to an experimental validation of our MPS-based approach to imputation, demonstrating first the ability of MPSTime to model complex joint distributions, and second, leveraging these learned distributions to perform an accurate imputation of missing values.
We begin by validating our approach on a relatively simple joint distribution of trendy sinusoids.
By systematically varying the parameters of our synthetic data generative model, we then show that a sufficiently expressive MPS can capture increasingly complex joint distributions of time-series data.
Moving beyond synthetic examples, we apply our approach to impute missing values in three real-world datasets with unknown generative processes.

In order to demonstrate the robustness of our MPS-based approach under varying degrees of data loss, we varied the proportion of missing data points in each imputation task by increasing the percentage of missing data from 5\% to 95\% of all samples.
In this work, we specifically focused on the contiguous missing data setting, where blocks of consecutive time-series observations were removed from test set instances.
Note that if missing data are unmeasured future values, such an imputation task could be considered a type of forecasting problem.
Imputation performance was evaluated in the original data domain using the mean absolute error (MAE) defined as:
\begin{equation} \label{eq:MAE}
    \text{MAE} = \frac{1}{n}\sum_{i=1}^n \lvert y_i - \hat{y}_i \rvert\,,
\end{equation}
where $y_i$ is the actual missing value for the $i$-th data point, $\hat{y}_i$ is the imputed value for the $i$-th data point, and $n$ is the number of missing values.

We compared the imputation performance of our approach to several benchmarks, each selected to encompass a diverse range of approaches in the time-series imputation literature including:
(i) 1-Nearest Neighbor Imputation (1-NNI); (ii) Centroid Decomposition Recovery (CDRec);
(iii) Bidirectional Recurrent Imputation for Time Series (BRITS-I); and
(iv) Conditional Score-based Diffusion Imputation (CSDI).

The simple benchmark, 1-NNI \cite{Beretta2016:NearestNeighborImputation}, imputes missing data points by identifying the nearest neighbor in the training set (based on a Euclidean distance of observed samples), then substitutes the missing values with those from the neighboring time series.
CDRec \cite{khayatiMemoryefficientCentroidDecomposition2014} decomposes a time-series matrix (i.e., an $N \times T$ matrix of $N$ time-series instances, each with fixed length $T$) into centroid patterns that capture recurring temporal behaviors, enabling the recovery of missing values through pattern-based reconstruction. 
The generative modeling benchmark, CSDI \cite{Tashiro2021:CSDIConditionalScorebased} uses a neural network-based diffusion model to gradually convert random noise into plausible imputed data points that are consistent with the data distribution conditioned on observed values.
Finally, BRITS-I \cite{caoBRITSBidirectionalRecurrent2018} is a popular deep-learning algorithm which uses recurrent neural networks (RNNs) to model temporal dependencies in both forward and backward directions.
While CDRec and CSDI were initially intended for multivariate time-series imputation, CDRec is one of the best performing algorithms in `total blackout' scenarios \cite{khayati_mind_2020}, and CSDI is the best performing generative model-based imputation algorithm \cite{Wang2024:DeepLearningMultivariate, du2023pypots}, which most closely matches the distribution-learning approach of MPSTime.

\begin{figure*}
    \centering
    \includegraphics[width=1.0\linewidth]{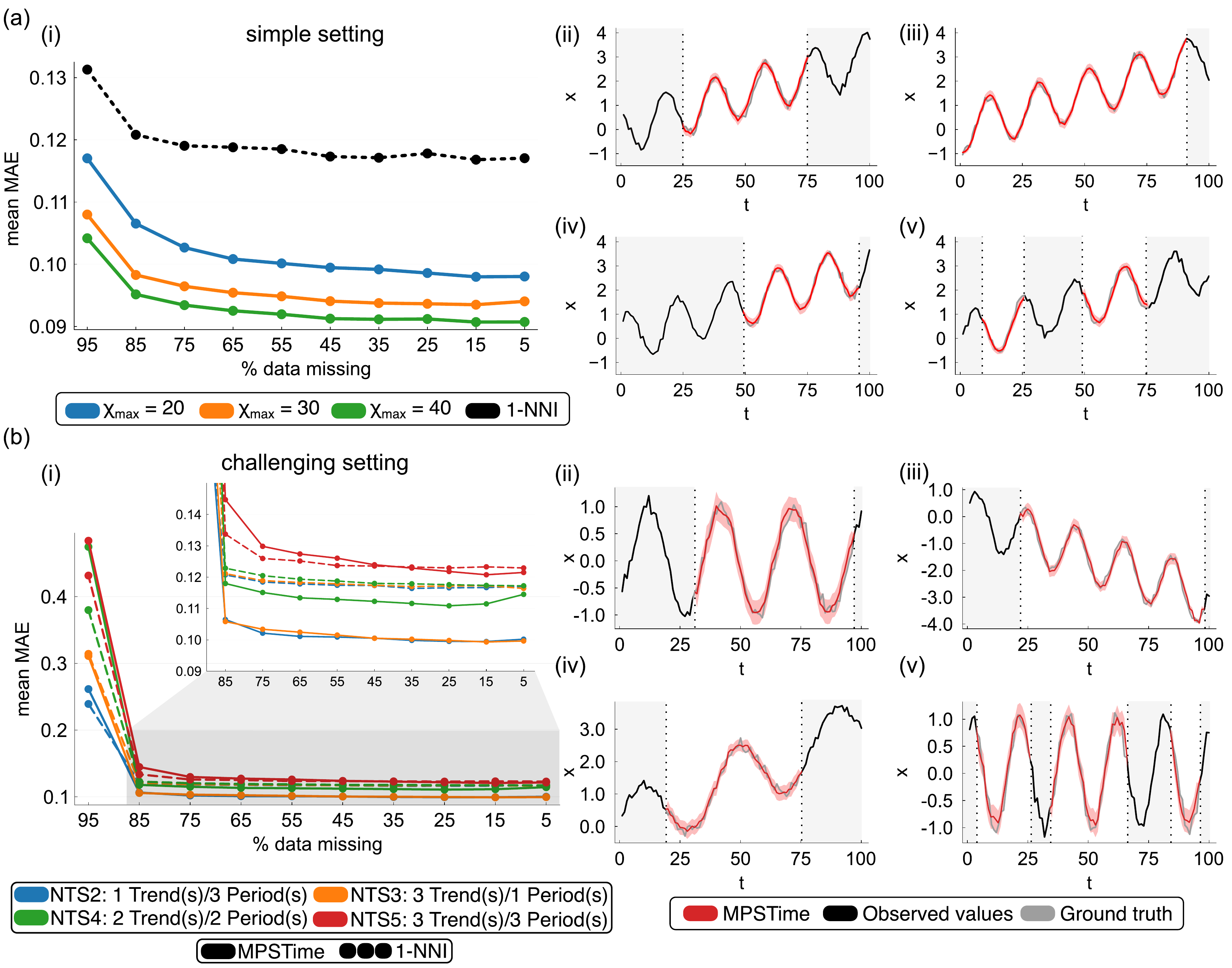}
    \caption{\textbf{Matrix-Product State (MPS)-based time-series imputation of synthesized datasets.}
    We compare MPSTime (solid line) against the 1-nearest-neighbor imputation (1-NNI) benchmark (dashed black line) on synthetic datasets of phase-randomized, noisy trendy sinusoids (NTS) generated by the model defined in Eq.~\eqref{eq:noisy-trendy-sinusoid}.
    The test set mean absolute error (MAE) between the imputed values and ground-truth (unobserved) values is reported across varying percentages of missing data. 
    \textbf{(a)} \textit{Simple setting:} MPSTime with three different bond dimensions $\chi_{\rm{max}} = 20, 30, 40$ (blue, orange, and green, respectively) is trained on a dataset exhibiting limited dynamical variation and a correspondingly simpler joint distribution with fixed trend $m$, fixed period $\tau$, and noise level $\sigma = 0.1$.
    \textbf{(b)} \textit{Challenging setting:}
    MPSTime with parameters determined by hyperparameter tuning trained on four datasets NTS2--5 (blue, orange, green, red, respectively) that exhibit richer dynamical variability and increasingly complex underlying joint distributions, constructed by varying the trends $m$ and/or periods $\tau$ of noise-corrupted, phase-randomized sinusoids with $\sigma = 0.1$, for the model defined in Eq.~\eqref{eq:noisy-trendy-sinusoid}.
    The upper-right inset zooms in on missing data percentages from $5\%$ to $85\%$ to highlight performance differences across the datasets.
    In both (a) and (b), we show representative time-series examples of the MPSTime-imputed values on unseen (`test') time series generated from the same model [NTS1 dataset with $\chi_{\rm max} = 40$ for (a) and NTS5 dataset with $\chi_{\rm max} = 160$ for (b)] in the panels (ii)--(v).
    The red shading in representative time-series examples indicates the imputation uncertainty, which here is quantified using the weighted median absolute deviation (WMAD).}
    \label{fig:interp-complex-synth}
\end{figure*}

\subsubsection{Synthetic time series}
\label{sec:imputation-synthetic}

In this section we investigate the ability of MPSTime to impute missing values from time-series data distributions of noisy trendy sinusoids (NTS) with incrementally increasing diversity, generated through different choices of the parameters in the common generative model in Eq.~\eqref{eq:noisy-trendy-sinusoid}. 
For our simplest imputation setting, NTS1, we trained MPS with three different fixed bond dimensions, $\chi_{\rm max} = 20, 30, 40$ for ten sweeps.
The learning rate $\eta$ and physical dimension $d$ were both determined by hyperparameter tuning with search ranges summarized in Appendix~\ref{appendix:imputation-hyperparm-opt}.
For a given percentage of missing data in each unseen test instance, we masked a random contiguous block of samples, and the missing segment was imputed using MPSTime.

Our first key result is presented in Fig.~\ref{fig:interp-complex-synth}(a), and shows the imputation error, reported as the mean absolute error (MAE), versus the percentage of missing data.
The relatively low MAE values for all bond dimensions $\chi_{\rm max}$ demonstrate that an MPS can effectively learn the distribution underlying a time-series dataset, and using the learned distribution, accurately impute missing values. 
This is reflected in the improvement in imputation performance as the bond dimension was increased, as more expressive MPS are able to capture more correlations, and therefore posses the capacity to infer missing values more accurately.
For all $\chi_{max}$ values, imputation performance improved as the number of observed data points increased, this is expected, as less missing data provides the MPS with additional conditioning information.
This improvement tends to saturate for $< 45\%$ missing data, as once the MPS gains enough information from the observed points to infer the phase of the signal to a sufficient degree of accuracy, aside from noise, it can then approximate all other values to a similar level of accuracy.

We observed a similar saturation in performance for the 1-NNI benchmark (dashed black curve), however, occurring at a higher MAE.
This is because only a small number of observations are required to fix the nearest training example, after which 1-NNI repeatedly returns the same example, and therefore cannot reduce its error below the residual train--test instance mismatch (within noise).

To demonstrate the imputation performance of MPSTime visually, four selected examples of the MPSTime-imputed time series for varying percentages of missing data are shown in Figs~\ref{fig:interp-complex-synth}(a)(ii)-(v). Fig.~\ref{fig:interp-complex-synth}(a)(v) shows an extension of the imputation task to a multi-window setting (note that for simplicity, multi-window imputation data is not included in Fig.~\ref{fig:interp-complex-synth}(a)(i)], however we expect similar imputation performance to the single-window setting).

All three MPS models with bond dimensions $\chi_{\rm max} = 20, 30, 40$, shown as the blue, orange and green curves, respectively, in Fig.~\ref{fig:interp-complex-synth}(a) consistently outperformed the 1-NNI benchmark (dashed black curve) across all missing data percentages.
The imputation results here indicate that, for small to moderate bond dimensions of $\chi_{\rm max}$ = 20-40, the MPS is sufficiently expressive to learn the joint probability distribution of the noisy sinusoid time series with fixed trend and unknown phase, and can accurately impute missing values, achieving MAE below 0.1 even when 85\% of time-series data points are missing. 

Having demonstrated that MPSTime can successfully encode the joint distribution underlying a relatively simple noisy periodic dataset, we next aimed to test its ability to model a more complex range of oscillatory dynamics.
To achieve this, we investigated four more complex versions of the underlying process in Eq.~\eqref{eq:noisy-trendy-sinusoid} (NTS2--5) by allowing the trend $m$ and/or the period $\tau$ to vary while holding the noise scale fixed at $\sigma = 0.1$.
The optimal values for $\chi_{\rm max}$, $\eta$ and $d$ were determined via a random search, as described in Appendix~\ref{appendix:imputation-hyperparm-opt}.

Figure~\ref{fig:interp-complex-synth}(b) shows the mean imputation error versus the percentage of data missing across the four datasets (NTS2--5).
We observe that the MAE for 1-NNI and MPSTime has already approximately saturated when the proportion of missing data has dropped to around $85\%$.
This is similar to the simple dataset (NTS1) in Fig.~\ref{fig:interp-complex-synth}(a) where, once a sufficient amount of the data is present, the projected MPS can accurately infer the relevant unknown properties of the process (e.g., the phase $\psi$, period $T$, and/or trend $m$), and accurately impute missing values.

Across the challenging datasets in Fig.~\ref{fig:interp-complex-synth}(b)(i), MPSTime imputation performance (solid curves) decreased with dataset diversity, reflecting the increasingly difficult challenge of modeling a broader range of temporal correlation structures.
As shown in Fig.~\ref{fig:interp-complex-synth}(b)(i) MPSTime performed best on the two datasets with least diversity, NTS2 with three possible periods and fixed trend (solid blue curve) and NTS3 with three possible trends and fixed period (solid orange curve).
With the exception of NTS5 (solid red curve), MPSTime outperformed the 1-NNI baseline (dashed curve for each dataset) for almost all percentages of missing data.
Our hyperparameter search for NTS5 consistently selected the largest possible bond dimension $\chi_{\rm max} = 160$, suggesting that further increasing the expressive capacity of the MPS through larger $\chi_{\rm max}$ could yield further improvements, particularly for $> 45\%$ data missing.
As we will see in the next section, the nearest-neighbor benchmark is also unexpectedly effective at imputing complex univariate data, so achieving parity with our benchmark is not an indicator of poor performance.

Four examples depicting the imputation performance for varying percentages of missing data in the most challenging setting (NTS5), with three distinct periods, trends, and random phase, are shown in Figs~\ref{fig:interp-complex-synth}(b)(ii)--(v).
The first three subplots, Figs~\ref{fig:interp-complex-synth}(b)(ii)--(iv), correspond directly to the analysis presented in Fig.~\ref{fig:interp-complex-synth}(b)(i), where performance of MPSTime is evaluated for a single contiguous block of missing data positioned at various locations within the time-series instances.
The last subplot, Fig.~\ref{fig:interp-complex-synth}(b)(v), showcases MPSTime's imputation behavior when multiple missing periods are present.
While this scenario is not explicitly considered in the analysis presented in Fig.~\ref{fig:interp-complex-synth}(b)(i), we include it here as visually compelling example to highlight the robustness and flexibility of MPSTime across diverse missing-data settings.

In general, the results presented in Figs~\ref{fig:interp-complex-synth}(a),(b) demonstrate that the MPS framework, when given sufficient training data, can learn the joint probability distribution of a dynamical process to a degree such that it can accurately infer the probabilistic model underlying trending sinusoidal functions.
The successful imputation of the more complicated dataset, NTS5, indicates that the MPS is sufficiently expressive to encode the generative model of a process with varying period, trend, and phase.
Together, these results support the usefulness of MPS as the basis for developing novel algorithms for analyzing time series.

\subsubsection{Real-world time series}
\label{sec:real-world-imputation-results}

Having demonstrated that MPSTime accurately imputes missing values from unseen time series, we now investigate its performance on real-world time-series datasets from three domains of application: (i) medicine (the `ECG' dataset); (ii) industry (the `Power Demand' dataset); and (iii) astronomy (the `Astronomy' dataset).
Further details about these datasets appear in Sec.~\ref{sec:real-world-dataset-details}.
To avoid bias from any specific train--test split, we used a cross-validation strategy with 30 resampled train--test splits for each dataset \cite{Bagnall2017:GreatTimeSeries}.
We tuned a separate MPS on each train--test split using 5-fold cross validation on the training set, then evaluated imputation error on unseen test instances.
Hyperparameter optimization was performed using a random search variant, with tuning ranges provided in Appendix~\ref{appendix:imputation-hyperparm-opt}.
For each test instance, we evaluated imputation performance by removing a contiguous segment of values, equivalent to a percentage of the total time-series length.
As in Sec.~\ref{sec:imputation-synthetic}, we examined missing data scenarios from 5\% to 95\%, in 10 percentage point increments. 
At each percentage level, we randomly varied the starting position of the missing segment to prevent bias from temporal segments with potentially more predictable patterns than others.
Using the same train--test splits and missing segment locations, we evaluated the imputation performance of the four benchmark algorithms described in Sec.~\ref{sec:imputation-experiments}: 1-nearest-neighbor imputation (1-NNI), a generative model-based imputation approach (CSDI), a recurrent neural network (BRITS-I), and a matrix completion method (CDRec).

\begin{figure*}
    \centering
    \includegraphics[width=1.0\linewidth]{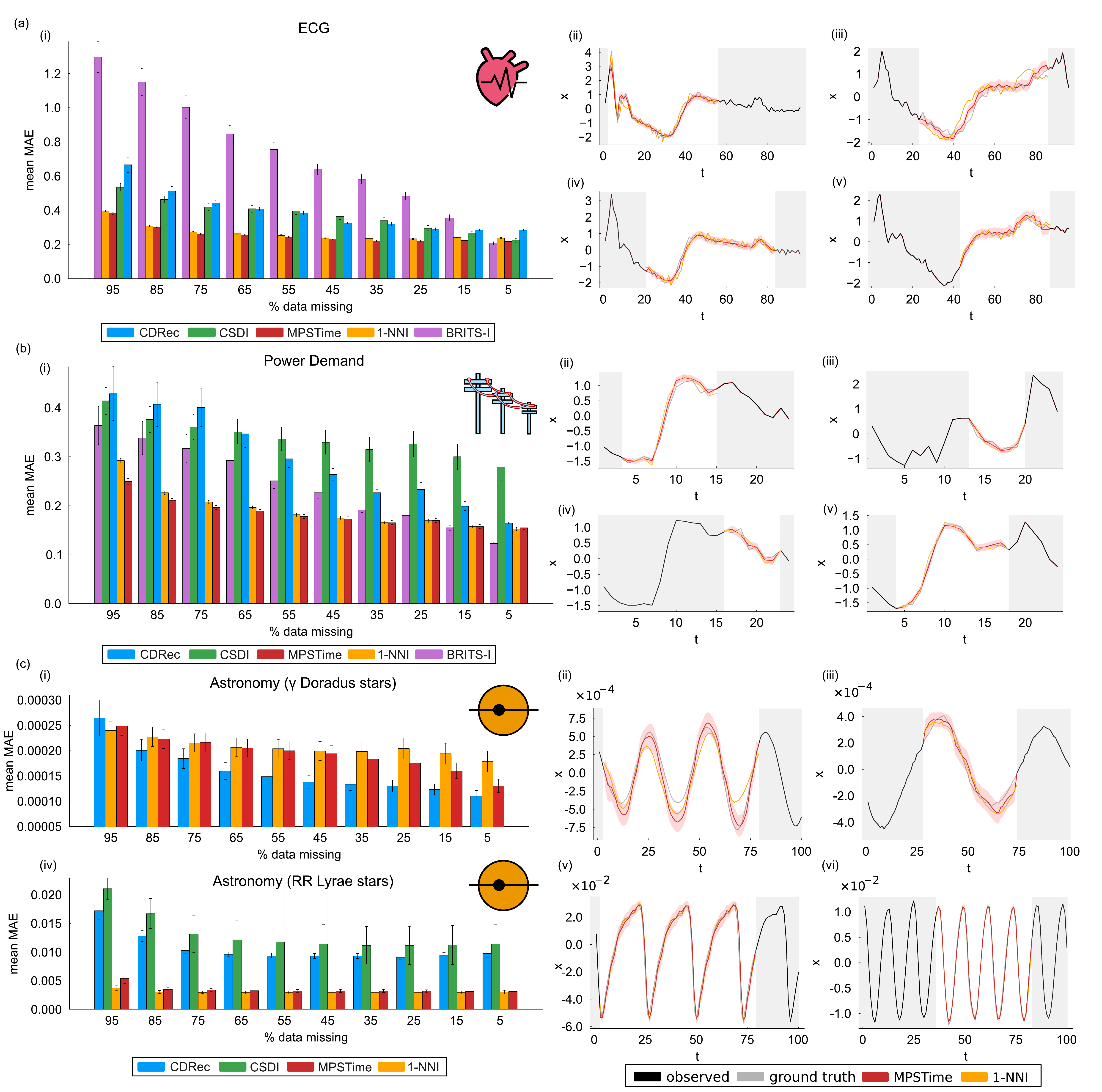}
    \caption{\textbf{MPSTime exhibits competitive (and often superior) performance to specialist benchmark algorithms for time-series imputation on real-world datasets.}
    Imputation error, reported as the Mean Absolute Error (MAE) and 95\% CI across 30 test folds is shown as a function of percentage data missing for each dataset in panels \textbf{(i)}: (a)(i) ECG, (b)(i) Power Demand, (c)(i, iv) Astronomy (see Sec.~\ref{sec:real-world-dataset-details}).
    MPSTime (red) is compared against four benchmarks: CDRec (blue), CSDI (green), 1-NNI (gold), and BRITS-I (purple).
    For Astronomy, the imputation performance on (c)(i) $\gamma$ Doradus and (c)(iv) RR Lyrae is plotted separately due to differing MAE scales.
    Benchmarks with substantially higher errors for $\gamma$ Doradus stars (BRITS-I and CSDI) and RR Lyrae stars (BRITS-I) were excluded from the Astronomy panels to maintain visual clarity among the top-performing algorithms.
    Panels (ii)--(v) for ECG and Power Demand show representative MPSTime imputations (solid red line), uncertainty due to encoding error (shaded ribbons), and the corresponding behavior of the 1-nearest neighbor (1-NNI) benchmark (solid gold line).
    For Astronomy, the panels (c)(ii, iii) and (c)(v, vi) show selected examples of the best performing MPSTime imputations for $\gamma$ Doradus stars and RR Lyrae stars, respectively. 
    In each imputation example, shaded regions denote observed segments, and transparent regions indicate missing (unobserved) data blocks.
    Ground-truth (unobserved) time-series values (gray line) are overlaid for reference.} 
\label{fig:empirical-imputation-results}
\end{figure*}

The imputation performance of MPSTime relative to the four benchmark algorithms is shown in Figs~\ref{fig:empirical-imputation-results}(a)--(c) for the ECG, Power Demand, and Astronomy datasets, respectively.
Starting with the ECG dataset, Fig.~\ref{fig:empirical-imputation-results}(a)(i) shows the mean MAE versus the percentage of missing data.
On this dataset, MPSTime exhibited superior performance to other benchmark algorithms across the full range of missing data percentages.
As with the synthetic datasets in Sec.~\ref{sec:imputation-synthetic}, the performance of MPSTime (red bar) tended to saturate with increasing missing data, in this case around the point of $< 65\%$ data missing. 
As above, this suggests that the MPS is able to `lock on' to the signal once a sufficient amount of information is present, beyond which there are minimal performance gains.
Figures~\ref{fig:empirical-imputation-results}(a)(ii)--(v) show selected examples of imputation performed by MPSTime (red curve), relative to the 1-NNI benchmark (gold curve) for different window sizes (i.e., $\%$ missing) and positions in the ECG dataset.
These plots show that, despite highly nonlinear and complex time-series patterns (relative to the synthetic datasets analyzed in Sec.~\ref{sec:imputation-synthetic} above), MPSTime can nevertheless encode an accurate approximation to the underlying data distribution, allowing it to accurately impute across missing periods of data from only a small fraction of measured data.

MPSTime also exhibited superior imputation performance on the Power Demand dataset for almost all percentages of data missing, as shown in Fig.~\ref{fig:empirical-imputation-results}(b)(i).
This dataset contains time series of fewer samples ($24$ samples each), compared to ECG ($96$ samples each), and MPSTime (red) exhibited a similar saturation in performance, this time at around $< 55 \%$ data missing, again indicating that the MPS framework can learn the complex patterns in time series with relatively few conditioning points.
Among the benchmark algorithms, 1-NNI (gold) demonstrated the strongest performance across most fractions of missing data.
Other benchmark imputation methods, such as CDRec (blue) and BRITS-I (purple), exhibited competitive performance with MPSTime at low \% data missing (e.g., at $5\%$), but degraded significantly when larger fractions of the data were missing.
Figures~\ref{fig:empirical-imputation-results}(b)(ii)--(v) show selected examples of imputation in the Power Demand dataset with different configurations of missing data, visually demonstrating the ability of MPSTime (red) to accurately impute complex nonlinear patterns.

Finally, imputation performance on the Astronomy dataset is plotted in Fig.~\ref{fig:empirical-imputation-results}(c), where the mean absolute error (MAE) is shown separately for the two classes, (i) $\gamma$ Doradus stars and (iv) RR Lyrae stars, due to several orders of magnitude difference in their respective data and error scales.
For $\gamma$ Doradus stars in Fig.~\ref{fig:empirical-imputation-results}(c)(i), the matrix decomposition benchmark CDRec (blue) outperformed both MPSTime (red) and 1-NNI (gold) for $< 85 \%$ data missing.
Unlike the previous examples, the imputation performance of MPSTime continued to improve as additional data points were made available.
The two neural network-based benchmarks, BRITS-I and CSDI, were excluded from the analysis presented in Fig.~\ref{fig:empirical-imputation-results}(c)(i) due to substantially larger errors (MAE $> 0.007$ for BRITS-I and MAE $> 0.004$ for CSDI, for all \% missing data).
Selected examples of the best performing MPSTime imputations (red) are shown in Figs~\ref{fig:empirical-imputation-results}(c)(ii),(iii).

For RR Lyrae in Fig.~\ref{fig:empirical-imputation-results}(c)(iv), MPSTime (red) and 1-NNI (gold) performed similarly for $< 75 \%$ missing data, with both outperforming the specialist algorithms for imputation across the full \% missing range.
Notably, for $< 95 \%$ data missing, the performance of both MPSTime and 1-NNI did not show a continual improvement or saturation effect after a fixed number of data points were known.
This is not unexpected due to the highly periodic nature of RR Lyrae star time series, as shown in Figs~\ref{fig:empirical-imputation-results}(c)(v),(vi), where, similar to the NTS datasets in Sec.~\ref{sec:imputation-synthetic}, MPSTime is able to infer the phase of the signal with relatively few points, beyond which there are no substantial performance gains.
This performance saturation closely matches the expected behavior of 1-NNI (gold) on a highly regular and periodic dataset where matching to a near-identical example with similar phase from the training set is likely.

The imputation results presented for real-world datasets in Fig.~\ref{fig:empirical-imputation-results} show that MPSTime can learn the joint probability distributions that underlie real-world time-series datasets with sufficient precision to perform accurate imputation. 
A notable aspect of the MPSTime performance was that, for the ECG and Power Demand datasets, the joint probability distribution was learned with a sufficient degree of accuracy that when $<85\%$ of the data was missing, MPSTime could perform imputation with a MAE value of $<0.3$ for ECG and $<0.2$ for Power Demand.
For these datasets, MPSTime almost always outperformed all other benchmarks, including specialist algorithms for time-series imputation.
Although MPSTime did not outperform in the Astronomy dataset, overall, its performance was among the best in all the imputation tests we performed. 
MPSTime therefore offers a promising approach for performing imputation on a diverse range of real-world time-series datasets for which the underlying generative process may be unknown.

\subsection{Time-series classification}
\label{section:classification-results}

Having shown that MPSTime can accurately impute missing values across multiple challenging scenarios involving synthetic and real-world data, we now demonstrate its ability to infer data classes through a series of time-series classification (TSC) tasks.
To underscore the inherent versatility of MPSTime in addressing diverse ML objectives within a unified framework, we constructed classification tasks from the same open real-world datasets (Sec.~\ref{sec:time-series-datasets}) analyzed in the imputation experiments above.

For each TSC task, we compare the performance of MPSTime to three benchmarks, each selected to represent a distinct algorithmic paradigm within the TSC literature.
Our selection of benchmarks comprise:
(i) Nearest Neighbor with Dynamic Time Warping (1-NN-DTW) \cite{Bagnall2017:GreatTimeSeries};
(ii) InceptionTime \cite{Fawaz2020:InceptionTimeFindingAlexNet}; and
(iii) HIVE-COTE V2.0 (HC2) \cite{Middlehurst2021:HIVECOTENewMeta}.
The classical elastic distance-based classifier, 1-NN-DTW, is a standard benchmark against which many new algorithms are compared, given its strong performance across various TSC problems on the UCR repository \cite{Bagnall2017:GreatTimeSeries}.
At the time of the analysis, the specialized deep neural-network benchmark, InceptionTime, was considered a state-of-the-art deep learning algorithm based on recent benchmarking \cite{Middlehurst2024:BakeReduxReview}.
Finally, HC2 is a meta-ensemble of classifiers, each built on different time-series data representations (e.g., shapelets, bag-of-words based dictionaries, among others).
Recent results placed HC2 as the top-performing algorithm on the UCR TSC repository \cite{Middlehurst2024:BakeReduxReview}.
Further discussion about these benchmark algorithms for TSC is in Appendix~\ref{appendix:tsc-baselines}.
We emphasize that our selection of HC2 and InceptionTime as benchmarks -- both highly expressive and computationally intensive -- establishes an especially challenging setting in which to compare the performance of MPSTime.

We evaluated classification accuracy using cross-validation across 30 stratified train-test resamples, as per \citet{Middlehurst2024:BakeReduxReview}.
For the ECG and Power Demand datasets, we used the same train--test split permutations from the UCR repository \cite{Dau2019:UCRTimeSeries}, matching those used for the published 30-fold mean accuracy results. 
The Astronomy dataset was newly derived for this study from the \texttt{KeplerLightCurves} open dataset (see Sec.~\ref{sec:real-world-dataset-details}).
As published results are not available for the benchmarks, we evaluated reference implementations of 1-NN-DTW, InceptionTime and HC2 on the Astronomy dataset, following the same 30-fold train-test resample approach outlined above.
Training details, including hyperparameter settings and specific implementations for the benchmarks are in Appendix~\ref{appendix:tsc-baselines}.

\begin{figure}
    \centering
    \includegraphics[width=1.0\linewidth]{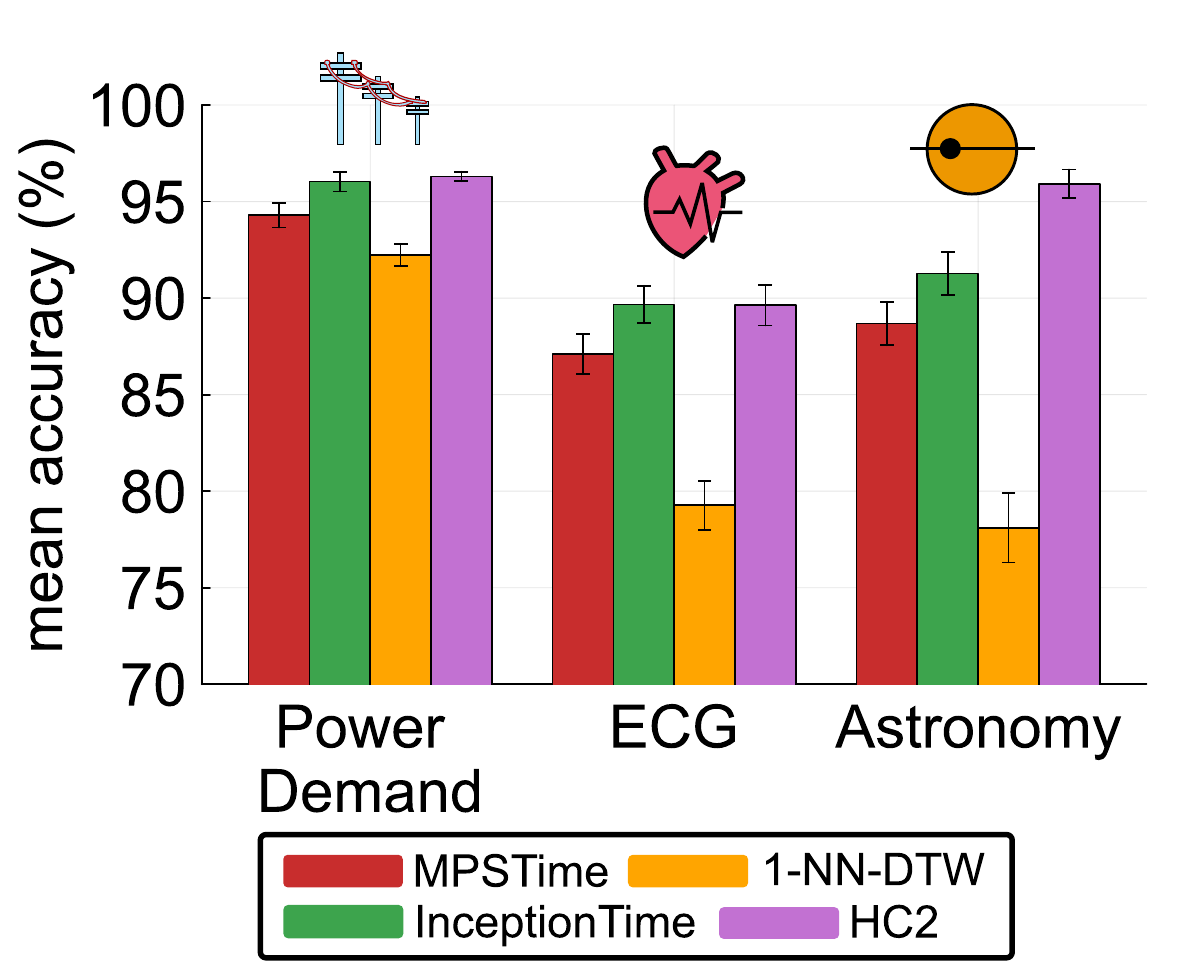}
    \caption{\textbf{MPSTime outperforms a classic time-series classification benchmark (1-NN-DTW) and is competitive with modern state-of-the-art classification algorithms: the neural-network-based InceptionTime and the meta-ensemble HIVE-COTE V2.0 (HC2).}
    The 30-fold mean classification accuracy and 95\% confidence interval (CI) are shown for each of the three empirical datasets: Power Demand, ECG, and Astronomy (see Sec.~\ref{sec:time-series-datasets}).
    We compare our MPS classifier, MPSTime (red), with three benchmarks: one-nearest neighbor with Dynamic Time-Warping (1-NN-DTW) (gold), InceptionTime (green), and HIVE-COTE V2.0 (HC2) (purple).
    }
    \label{fig:classification-barplots}
\end{figure}

The classification results, shown in Fig.~\ref{fig:classification-barplots}, highlight the strong performance of MPSTime, which achieves TSC accuracies competitive with state-of-the-art TSC algorithms on our benchmarks.
Notably, for both ECG and Power Demand datasets, MPSTime achieves accuracies that are competitive with top-performing deep learning and ensemble methods, InceptionTime and HC2, respectively, while outperforming the classic 1-NN-DTW benchmark on these datasets.
On ECG, our preliminary findings (see Appendix.~\ref{appendix:tsc-training-times-comparison}) also suggest that MPSTime achieves a favorable accuracy/train-time trade-off relative to the benchmarks.
Finally, on the Astronomy dataset, we observed the greatest variability in performance differences across methods, with MPSTime achieving a mean accuracy of $88.7\%$ ($95\%$ CI: $87.6\% - 89.8\%$), which was lower than both InceptionTime, $91.3\%$ ($95\%$ CI: $90.2\% - 92.4\%$), and HC2, $95.9\%$ ($95\%$ CI: $95.2\% - 96.6\%$).

In summary, our findings indicate that an initial adaption of MPSTime to classification, without sophisticated hyperparameter tuning procedures, nevertheless achieves competitive performance (and relatively efficient training cost) on real-world time-series datasets when compared to state-of-the-art methods.
These results demonstrate both the versatility of our framework in tackling multiple time-series ML objectives (i.e., both imputation and classification) and reveal significant opportunities for improving the performance of MPSTime through further specialization.

\section{MPS interpretability}
\label{sec:interpretability}

\begin{figure*}
    \centering
    \includegraphics[width=1.0\linewidth]{Figures/Results/fig_7_synthetic_datagen.pdf}
    \caption{\textbf{Sampling MPS joint probability distributions over time series yields synthetic examples that closely resemble the training data.}
    At each time-point, the time-marginalized kernel density estimate (KDE) of time-series amplitudes is visualized as a heat map, and is overlaid with individual time-series instances from the entire training set (white traces).
    The color intensity within each heatmap indicates the time-series amplitude density relative to the maximum at each time-point.
    For each dataset -- \textbf{(a)} ECG, \textbf{(b)} Power Demand, and \textbf{(c)} the most complex Noisy, Trendy Sinusoid dataset (NTS5) from Sec.~\ref{sec:imputation-synthetic} with 3 periods and 3 trends -- we show an upper and a lower panel: in subplot \textbf{(i)}, we show the ground-truth training data as white traces, with the KDE computed from the full training set across all data classes.
    Subplot \textbf{(ii)} shows the synthetic data generated by sampling from a sufficiently expressive $\chi_{\rm max} = 80$, $d = 18$ MPS trained on the same dataset.
    Using inverse transform sampling (see Appendix~\ref{appendix:inverse-transform-sampling}), we generate each time-series amplitude conditionally from the MPS-encoded distribution, and depict representative examples of the resulting trajectories with white traces. 
    }
    \label{fig:mps_vs_data_dist}
\end{figure*}

A compelling justification for using matrix-product states to infer joint distributions directly from time-series data lies in their inherent interpretability, which sets them apart from `black-box' approaches.
By inferring a closed-form approximation of the joint distribution, MPSTime permits analysis of the underlying time-series distribution that would be computationally intractable for standard methods.
In this section, we showcase two aspects of interpretable MPS-based machine learning that are enabled by studying the single-site reduced density matrix (RDM) $\rho$ in Eq.~\eqref{eq:outer-joint}.
At its core, the single site RDM captures both the local statistical properties of the time-series at a given point.
Crucially, the RDM is easily extractable directly from the MPS, and contains all the information needed to sample from the approximated joint distribution, as well as analyze the short-range and long-range temporal correlation structures that it encodes.

First, in Sec.~\ref{sec:sampling-joints}, by comparing samples generated by the MPS to the training dataset, we show that MPSTime can successfully capture complex temporal correlations from a finite number of training instances.
This indicates that MPSTime can be used to generate new synthetic time-series examples from any given data class, an application that has broad implications \cite{zanchi_synthetic_2025, esteban_real-valued_2017, huangTimeDPLearningGenerate2025}.
Second, in Sec.~\ref{sec:conditional-entanglement}, we show how the complex correlations encoded by the MPS can be understood through the lens of entanglement entropy, extending fundamental concepts from quantum information theory -- traditionally applied to analyze MPS in quantum many-body physics -- to the analysis of temporal correlation structures learned by MPSTime.
Third, in Sec.~\ref{sec:mps-ood-extrapolation}, we study the ability of MPSTime to generalize learned data structures to new, out-of-distribution (OOD) contexts.

\subsection{Qualitatively evaluating the learned joint distribution}
\label{sec:sampling-joints}

In Sec.~\ref{sec:imputation-experiments}, we demonstrated that MPSTime accurately imputes missing values in partially observed time series by sampling from its learned joint distribution, conditioned on observed values.   
Here, we extend this analysis to examine the full joint distribution encoded by the MPS, providing insight into what temporal correlations and patterns MPSTime learns from the training data. 
Instead of conditioning on observed values, as in Sec.~\ref{sec:imputation-experiments}, we now generate entire time-series instances by sampling values directly from the unconditioned MPS.
For clarity, we refer to MPS-generated data as `trajectories', and real training data as `time-series instances'.
Our sampling procedure follows the natural temporal ordering of time series, beginning at the first MPS site ($t = 1$) and advancing sequentially until reaching the final site ($t = T$).
Each time, a random state is selected from the reduced density matrix (RDM) using inverse transform sampling, and the MPS is updated according to the sampled value.
This site-wise process of sampling and updating the MPS continues until generating a complete trajectory of $T$ samples.
By returning to $t = 1$ and sampling again, additional trajectories can be generated from the MPSTime-modeled joint distribution.
For implementation details of our inverse transform sampling approach, which is similar to recent work by \citet{Mossi2024:MatrixProductState}, see Appendix~\ref{appendix:inverse-transform-sampling}.

For our investigation, we focused on one synthetic dataset (NTS5) and two empirical datasets (Power Demand and ECG) from Sec.~\ref{sec:imputation-experiments}.
As with imputation, we discarded class-specific information for each dataset, where applicable, by treating each time-series instance as belonging to the same data class.
For the purposes of qualitative comparison, we selected values of $\chi_{\rm max}$ and $d$ that ensured the MPS was sufficiently expressive to model the essential temporal correlations (i.e., larger $\chi_{\rm max}$), while producing trajectories from the encoded generative distribution with small encoding error (i.e., larger $d$).
To achieve this, we trained a $\chi_{\rm max} = 80$, $d = 18$ MPS for all three datasets.
From each MPS, we then generated the same number of trajectories as there were time-series instances in the training set, allowing for a direct comparison between realizations from the various modeled and original data distributions.

In Fig.~\ref{fig:mps_vs_data_dist} we qualitatively compare the synthetically generated trajectories from MPSTime (lower panels) to training instances from the ground-truth joint distribution (upper panels) for each of the three datasets.
In general, for each real and synthetic dataset pair, we observe that the lighter regions of the synthetic dataset KDEs (which indicate higher relative density) overlap closely with the typical amplitude ranges seen in the ground-truth datasets.
This correspondence between the relative densities of the synthetic and real time-series datasets at each time point suggests that MPSTime has accurately approximated the underlying distribution.
Inspecting the individual time series in the upper panels, shown as superimposed white traces, we observe that the MPS-generated trajectories are comparatively less smooth than those from the real (training) dataset, particularly for the longer ECG and NTS5 time series.
This effect is to be expected, as the encoding and approximation errors inherent in the MPS representation, as discussed in Appendix~\ref{appendix:enc-error}, limit the precision with which finer temporal features can be resolved.    

Across all three datasets -- ECG, Power Demand, and NTS5 -- MPSTime captures essential qualitative temporal features present in the training data.
For the ECG dataset in Fig.~\ref{fig:mps_vs_data_dist}(a), synthetic trajectories in (a-ii) replicate the characteristic cardiac cycle of the training data in (a-i), including the pronounced QRS complex (initial peak) near $t = 1$ and the more subtle secondary T wave near $t = 80$.

For the Power Demand dataset in Fig.~\ref{fig:mps_vs_data_dist}(b), the generated trajectories in Fig.~\ref{fig:mps_vs_data_dist}(b-ii) capture the two-peak structure and intervening dip present in Fig.~\ref{fig:mps_vs_data_dist}(b-i), mirroring the cyclic patterns of real power usage over the 24 hour period. 
Compared to the training data, the Power Demand trajectories are less diverse.
Notably, very few trajectories begin with values above $x = 0$ at $t = 1$ or reach values exceeding $x = 2$ around $t = 22$. 
These discrepancies may arise from outlier values being under-represented by the MPS approximation of the joint distribution, an effect that is especially pronounced for small datasets such as IPD with only $67$ training instances.
Further, the diversity of trajectories generated by our rejection-based sampling approach -- implemented to avoid sampling unphysical states due to artificial distribution broadening (see Appendix.~\ref{appendix:inverse-transform-sampling}) -- further reduces the likelihood of observing these rare outlier values.

Finally, for the NTS5 dataset in Fig.~\ref{fig:mps_vs_data_dist}(c), the training data in Fig.~\ref{fig:mps_vs_data_dist}(c-i) exhibits a particularly complex temporal structure arising from sinusoids with multiple trends, periods and randomized phases, as described in Sec.~\ref{sec:dataset-details-synthetic}.
By superimposing time-series, shown as white traces, the training distribution reveals three distinct bands, corresponding to the three possible slopes ($m = -3, 0, 3$ in Eq.~\eqref{eq:noisy-trendy-sinusoid}) of noisy trendy sinusoid.
The MPS-generated trajectories in Fig.~\ref{fig:mps_vs_data_dist}(c-ii) largely reproduce these qualitative features by capturing the multi-period oscillatory and trend behavior, as well as the general amplitude ranges.
However, the synthetic dataset shows comparatively less distinct band separation and a more diffuse relative density in the KDE heatmap, particularly for $t < 20$.
Once again, this is likely due to the MPS approximation and encoding error, which results in noisier data that blur the boundaries between bands, reducing the resolution of finer temporal features, and leading to a more diffuse density structure than its real counterpart in Fig.~\ref{fig:mps_vs_data_dist}(c-i).

Overall, our findings suggest that MPSTime effectively replicates the data distribution underlying a finite number of time-series by generating synthetic data that closely resemble training data in their essential qualitative features.
This provides evidence that MPSTime learns meaningful temporal correlations and structure from the time-series data.
It should be noted that while synthetic trajectories were generally less smooth than their real counterparts, this can be mitigated by increasing the physical dimension $d$ of the MPS (see Appendix~\ref{appendix:enc-error}), although doing so would incur higher computational costs. 
While we limit the scope of our investigation to a qualitative comparison of temporal features, future work could quantify distributional similarity through low dimensional projections, classifier-based discrimination, or metrics suited to high-dimensional spaces, among others \cite{bischoffPracticalGuideSamplebased2024}.

\begin{figure*}
    \centering
    \includegraphics[width=1.0\linewidth]{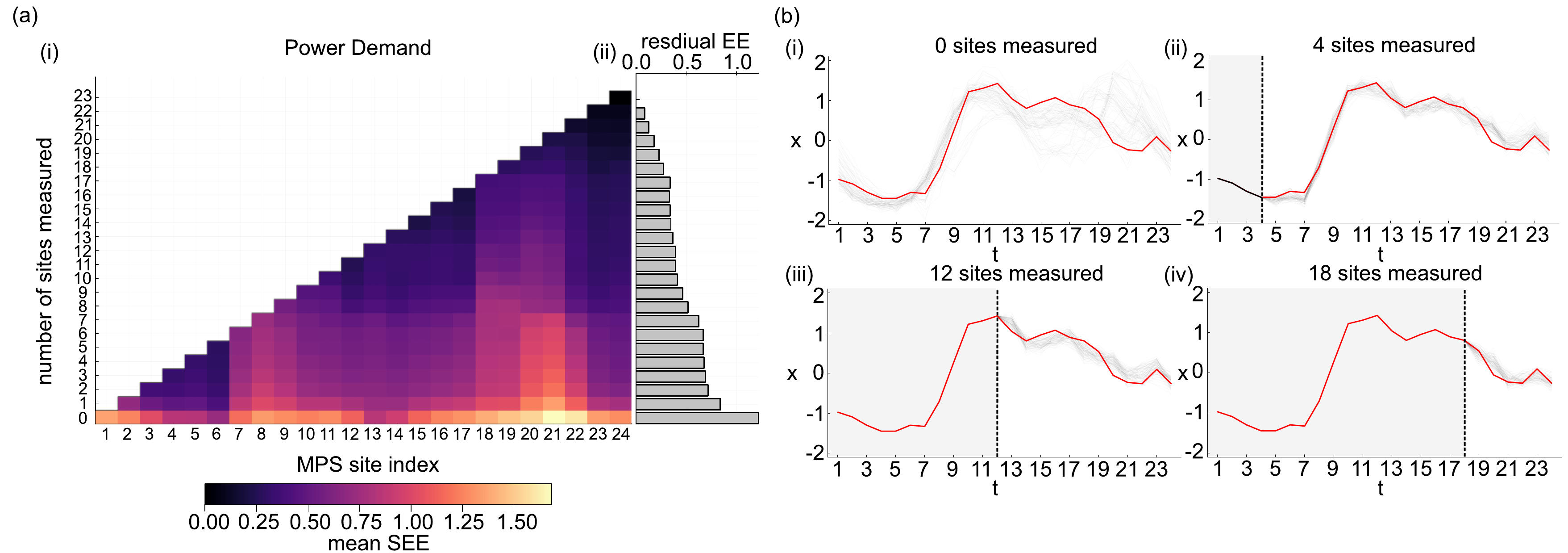}
    \caption{\textbf{Interpreting the joint probability distribution encoded by a trained MPS using the single-site entanglement entropy (SEE).}
    By sequentially updating a trained MPS ($d = 12$, $\chi_{\rm max} = 80$) with each additional measured site, the SEE reveals how the entanglement in the remaining unmeasured sites evolves.
    This SEE structure provides insight into the contextual dependencies captured by the MPS, which, as it is conditioned on each additional site, reflects an evolving conditional distribution. 
    Panel \textbf{(a)} shows the mean SEE, with magnitude indicated by the color intensity, across the Power Demand test set as a function of the number of sequentially measured sites (vertical axis), starting from the first site and progressing from left ($t = 1$) to right ($t = 24$).
    The accompanying bar plot depicts the residual entanglement entropy -- defined as the average unmeasured SEE -- for varying numbers of measured sites.
    Panel \textbf{(b)} shows representative time-series trajectories sampled from the MPS for varying numbers of measured sites.
    These trajectories demonstrate how the conditional distribution of time-series amplitudes evolves as increasingly many points are measured, for: (i) no sites measured; (ii) 4 sites measured; (iii) 12 sites measured; and (iv) 18 sites measured.
    The red curve in each panel depicts an unseen time-series instance from the test set for which time points lying to the left of the vertical dashed line (shaded window) have been used to condition the MPS.
    }
    \label{fig:mps_see_plot}
\end{figure*}

\subsection{Extracting conditional entanglement entropy from the MPS}
\label{sec:conditional-entanglement}

In a quantum mechanical system, the single-site entanglement entropy (SEE) quantifies the amount of entanglement shared between a single site and the remainder of the system \cite{Liu2021:EntanglementBasedFeatureExtraction}. 
When adapted to an MPS trained on a time-series dataset, the 
SEE determines the extent to which the time-series amplitude at a single site depends on the value of all other amplitudes (see Fig.~\ref{fig:conceptual-picture}(c)).
If the SEE is large, this implies strong correlations with the remainder of the MPS, i.e., determining the amplitudes at other sites (time points) can reduce uncertainty about the amplitude at the site under consideration.
On the other hand, a small SEE suggests the site behaves more independently, and therefore, determining the amplitudes of other sites does little to constrain its value through uncertainty reduction.

Examining how the SEE changes after conditioning on the time-series amplitude at one or more sites can therefore provide insight into the conditional relationships learned by the MPS.
As a concrete example, consider an unmeasured site $B$ with $\textrm{SEE}_{\rm before} > 0$.
We then `measure' another site $A$, projecting the MPS onto a subspace in which the local state at $A$ is fixed to the observed time-series value.
If this conditioning reduces the SEE at site $B$, i.e., $\textrm{SEE}_{\rm after} < \textrm{SEE}_{\rm before}$, this implies that $B$'s state is strongly influenced by the local state at $A$.
In other words, observing a particular time-series amplitude $x_A$ at site $A$ constrains the distribution of amplitudes $x_B$ can take at site $B$.
Extending this approach across multiple pairs or sets of sites can allow us to `map out' more complex correlation structures encoded by the MPS.

For an MPS site $A_i$, the SEE is defined as the von Neumann entropy of the single-site reduced density matrix~\cite{Hayashi2015:InformationQuantitiesQuantum}:
\begin{equation}
    S = - \sum_{k=1}^r
    \label{eq:entanglement-entropy} \lambda_k \ln{\lambda_k}\,,
\end{equation}
where $\lambda_k$ are the eigenvalues of the single-site reduced density matrix $\rho_i$ with rank $r$. 
The single-site density matrix is always positive semi-definite and therefore the non-zero eigenvalues $\lambda_k > 0$.

To showcase the insights facilitated by the SEE, we analyzed the same MPS trained on the Power Demand dataset in Sec.~\ref{sec:sampling-joints}.
We aimed to uncover the conditional dependencies learned by the MPS by observing how its entanglement structure evolves as increasingly many site-wise measurements are made.
We chose to perform measurements in a sequential order, starting at the first site and incrementing the number of measured sites by one.
With each additional measurement, we analyzed the SEE of the remaining unmeasured sites in order to isolate the influence of the measurement on the resulting entanglement structure.

In Fig.~\ref{fig:mps_see_plot}(a)(i), the conditional entanglement structure of the 24-site MPS is plotted as the average of the measurement-dependent SEE over the entire Power Demand test set (containing $1029$ time series).
As a proxy for the remaining entanglement encapsulated by the MPS after cumulative measurements, we also computed the residual entanglement entropy as the average SEE of the remaining unmeasured sites.
As expected, as more sites of the MPS are conditioned on known values, the SEE decreases. 
Interestingly, there appear to be key sites where, after they are conditioned, the SEE drops sharply.
In particular, measuring sites (corresponding to time points) $t = 1, 2, 8, 18$, result in comparatively large reductions in SEE, indicating that the values at these points are crucial in determining the value of other points within the MPS. 
Figure~\ref{fig:mps_see_plot}(b) shows the sampled trajectories of the joint probability distribution learned by the MPS, after conditioning the MPS on the first 0 to 18 sites, using known values from a selected training instance.
Moving from Figs~\ref{fig:mps_see_plot}(b)(i)--(iv), we observe a clear reduction in the fluctuations in the trajectory amplitudes as more sites are measured.
This reduction is particularly prominent between $t = 15$ and $t = 23$.
This indicates that the conditional SEE can be used as a proxy for estimating the uncertainty of unmeasured sites, conditioned on the measurement outcomes of other sites.
The results illustrated in Fig.~\ref{fig:mps_see_plot} highlight the utility of MPSTime, which can be used to compute the conditional trajectories and conditional SEE.

\begin{figure}
    \includegraphics[width=1.0\linewidth]{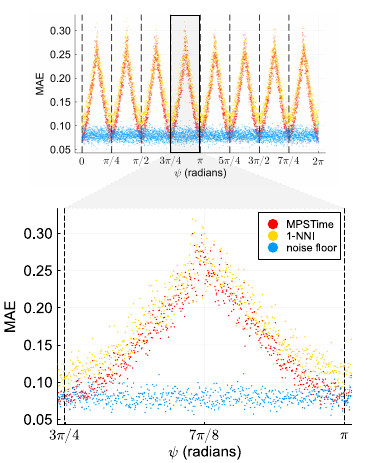}
    \caption{\textbf{Evaluating the out-of-distribution generalization of MPSTime on noisy trendy sinusoids (NTS) with unobserved phases.}
    MPSTime was trained on NTS (Eq.\eqref{eq:noisy-trendy-sinusoid}) with eight discrete phases (indicated by the vertical dashed lines in the upper plot).
    To test generalization, we swept the phase parameter $\psi$ in Eq.\eqref{eq:noisy-trendy-sinusoid} across the entire range $[0, 2\pi)$ and imputed 50\% missing data in unseen NTS with phase $\psi$. 
    The Mean Absolute Error (MAE) between imputed and original noisy ground-truth values is shown for each value of $\psi$. 
    The lower plot zooms into the range $[3\pi/4, \pi]$ to highlight the performance on intermediate phases.
    We compare the performance of MPSTime (red) with a 1-Nearest Neighbor imputation (1-NNI) baseline (gold), which cannot generalize beyond its training dataset.
    We also show a noise floor baseline (blue), which corresponds to the irreducible minimum MAE between noisy ground-truth values and a perfect reconstruction of the underlying noise-free signal (obtained using Eq.~\eqref{eq:noisy-trendy-sinusoid} with $\sigma = 0$ and the ground-truth phase $\psi$).
    }
    \label{fig:mps-phase-generalization}
\end{figure}

\subsection{Assessing MPS generalization to out-of-distribution instances}
\label{sec:mps-ood-extrapolation}

The ability of a trained model to usefully extrapolate beyond its training distribution is captured by its out-of-distribution (OOD) performance.
To determine the extent to which MPSTime generalizes usefully to OOD time series, we constructed a simple time-series imputation task that requires learning and extrapolating the dynamical rules of the observed data in order to demonstrate strong performance.
We focused on the noisy trendy sinusoid (NTS), Eq.~\eqref{eq:noisy-trendy-sinusoid}, with fixed trend ($m = 1.0$), period ($\tau = 30$), and Gaussian additive noise ($\sigma = 0.1$).
We constructed a restricted subset of training data ($1000$ NTS instances of length $T = 100$ samples each), which encompass a small set of sinusoidal phases $\psi$ that can only take one of eight possible values ($\psi = \frac{\pi k}{4}$, where $k = 0,1, \dots, 7$ is an integer).
MPSTime was trained on the restricted subset of data using a sufficiently large physical and maximum bond dimension, $d = 10$ and $\chi_{\rm max} = 60$, respectively.
We then assessed the ability of MPSTime to accurately impute noisy sinusoids with phases that differ from those in the training set by uniformly discretizing the phase interval $[0, 2\pi)$ into $5000$ points and generating an independent NTS test instance at each sampled phase.
In each test instance, we masked a random contiguous block of 50\% of the samples, and MPSTime was tasked with imputing the missing segment.
We evaluated MPSTime against a noise-floor baseline, representing the irreducible error obtained when replacing a missing NTS segment with its noise-free counterpart (generated at the same test phase from the same NTS model in Eq.~\eqref{eq:noisy-trendy-sinusoid} with $\sigma = 0$), and the same 1-NNI benchmark used above (in Sec.~\ref{sec:imputation-experiments}).

Imputation performance, reported as mean absolute error (MAE), is shown as a function of phase for all test instances in Fig.~\ref{fig:mps-phase-generalization}.
As expected, the noise floor baseline (shown in blue), yields a phase-independent MAE that captures the best achievable imputation performance in the absence of noise.
For the 1-NNI benchmark, test instances are matched to their nearest neighbor, typically corresponding to a sinusoid in the training set with the closest phase $\psi$ to a given test instance.
This behavior is reflected in the gold MAE$(\psi)$ curve in Fig.~\ref{fig:mps-phase-generalization}, which is minimal when the phase $\psi$ coincides with an example in the training set (vertical dashed lines), and increases with $\Delta \psi$ between a given test instance and the training set match.
MPSTime OOD performance (red in Fig.~\ref{fig:mps-phase-generalization}) is qualitatively similar, with the MAE peaking at midpoints between the two nearest phases in the training set, achieves consistently lower MAE values than 1-NNI across the phase range.
However, compared to 1-NNI, MPSTime has generally lower error, and matches the noise floor baseline at phases close to those contained in the training set (dotted vertical lines in Fig.~\ref{fig:mps-phase-generalization}), indicating that it has accurately imputed the underlying noise-free signal.
Relative to 1-NNI, which always matches to a single noisy instance in the training set, this points to the ability of MPSTime to better encapsulate the deterministic component of the signal. 
This is likely due to our MPS imputation algorithm choosing the median of the RDM probability distribution (see Appendix~\ref{appendix:enc-error}), which will tend to average over noise.

One way to interpret these findings is that MPSTime exhibits qualitatively similar behavior to 1-NNI through an MPS-based compression of the relevant temporal correlations contained in the training dataset.
This interpretation is consistent with previous work by \citet{martynEntanglementTensorNetworks2020} showing that instead of storing the entire training distribution of MNIST image data -- a task requiring prohibitively large entanglement -- an MPS learns a memory efficient low-entanglement representation, analogous to performing an efficient data compression.
We note that although the MPS in this example cannot generalize information from training data of selected phases to learn all other phases, this property could easily be introduced into the MPS geometry using a translationally invariant MPS \cite{perez-garciaMatrixProductState2007} which should, by construction, learn all phases from a single training phase.

\section{Discussion and Conclusion}
\label{sec:discussionAndConclusion}

In this work we have introduced MPSTime, a flexible algorithmic framework for learning complex time-series distributions using matrix-product states (MPS) that can be applied to a range of time-series machine-learning (ML) problems, including classification and imputation.
This capacity to tackle diverse time-series ML tasks within a unified probabilistic framework sets MPSTime apart from many conventional ML algorithms which focus on a single problem class in isolation (e.g., disjoint algorithms for time-series classification, imputation, etc.).
In addition, the ability to extract the learned temporal dependencies directly from the MPS (via the reduced density matrix) grants MPSTime a level of interpretability that is lacking in many other time-series ML approaches.
For the machine learning applications presented here, MPSTime required only moderate bond dimension ($\chi_{\rm max} = 20-160$) and physical dimension ($d = 2-15$), achieving a favorable balance between expressiveness and computational cost.
Taken together, our results support the MPS ansatz as an efficient and useful way to distill complex temporal dependencies contained in time-series data, paving the way for future work to extend our approach to tackle a wider range of time-series problems, including forecasting and synthetic data generation.

Building on a growing body of work applying MPS to machine learning, this work contributes several significant innovations to an emerging field.
For the first time, we develop an end-to-end framework for training MPS to approximate the joint distribution of continuous-valued sequential (time-series) data.
While generalized feature maps (e.g., Legendre and Fourier basis) have been used to encode continuous-valued inputs of at most four dimensions (i.e., an MPS of four sites) \cite{Meiburg2023:GenerativeLearningContinuous},
our approach scales to continuous-valued time-series data of up to 100 samples (i.e., an MPS of 100 sites) and beyond, while remaining practical to run, even on real-world datasets.
We introduce two complementary error reduction techniques -- leveraging the conditional median as a robust point estimate for time-series imputation, and a rejection sampling scheme that curbs the propagation of encoding-related errors with site-wise sampling.
Together, our methods overcome the inherent limitations of discrete encodings for continuous-valued time-series data, enabling accurate imputation and robust data generation on long sequences.
We anticipate that such algorithmic advancements will provide a foundation for MPS-based generative modeling applications.
We have made our implementations of the imputation, classification, MPS-analysis, and hyperparameter optimization algorithms used in this work publicly available in the comprehensively documented code package \textit{MPSTime}. 
Our implementation has been extensively optimized, and, unlike many publicly available time-series ML algorithms, does not require specialized knowledge or hardware to run. 

While our approach achieves performance comparable to state-of-the-art benchmarks on a range of time-series ML tasks, there is room for further improvement.
First, as a preliminary investigation, we focused on a small number of representative datasets to demonstrate the viability of MPS-based algorithms for time-series ML. 
While a comprehensive benchmark falls outside the scope of this study, it offers a clear road map for future investigations. 
A limitation of the fixed-length MPS geometry is that the current implementation of MPSTime can only handle time-series instances of fixed length.
Although this can be partially addressed through time-series windowing, such an approach neglects potentially crucial long-range temporal dependencies that extend beyond the chosen window length.
A promising avenue for future work is to explore translationally invariant MPS (TI-MPS) \cite{perez-garciaMatrixProductState2007} which have long been used in quantum many-body physics to approximate one-dimensional systems in the thermodynamic limit \cite{vidal_classical_2007}.
Owing to their flexible architecture, such TI-MPS can be naturally adapted to time series of varying lengths, opening up the possibility of time-series forecasting across arbitrary horizons.
Furthermore, for stationary time series, where the statistical properties are invariant to time shifts \cite{owensParameterInferenceNonstationary2024}, incorporating translational invariance into the MPS geometry could enhance out-of-distribution generalization.
As found in Sec.~\ref{sec:mps-ood-extrapolation}, the behavior of MPSTime can be approximately thought of as a nearest-neighbor-like efficient compression of the training distribution, which is desirable in many settings, but limits its out-of-distribution generalization.
One direction in which this functionality could be incorporated is through TI-MPS, which should generalize a stationary correlation structure to new time series; in the simple test case of noisy sinusoids with variable phase studied in Sec.~\ref{sec:mps-ood-extrapolation}, a TI-MPS should be able to accurately impute missing values in sinusoids with arbitrary phases after training on examples of only a single phase. 
This can have particular advantages for very large time-series datasets where nearest neighbor can become prohibitively slow.

There are many promising extensions of MPSTime to problem classes beyond classification and imputation, demonstrated here, including to anomaly detection \cite{schmidlAnomalyDetectionTime2022}, regression (of extrinsic variables from time series \cite{tanTimeSeriesExtrinsic2021}), forecasting, and synthetic data generation.
Such extensions are made relatively straightforward as they can be developed from a common underlying probabilistic model, MPSTime, similarly to other flexible learning algorithms for time-series modeling \cite{yangSurveyDiffusionModels2024}.
The application to synthetic time-series generation is a particularly promising direction given our results presented in Sec.~\ref{sec:sampling-joints}, where conditionally sampling from a trained MPS can yield an ensemble of time series that closely resemble the training dataset.
Future work should also assess the performance of MPSTime relative to a range of other flexible generative models for time-series data, including recent developments in this space using techniques including variational autoencoders (VAEs), generative adversarial networks (GANs), and denoising diffusion probabilistic models \cite{yangSurveyDiffusionModels2024}.

In conclusion, here we have demonstrated the usefulness of the MPS ansatz in efficiently capturing complex time-series data structures and introduced an open and extendable algorithmic implementation, MPSTime, that implements the method.
The high performance of MPSTime in classification and imputation settings along with its potential for interpretability make it a promising and flexible new approach to modeling complex time-series data, providing a common theoretical foundation on which algorithms for tackling a range of time-series problem classes can be developed.


\begin{acknowledgments}
This research was supported by the University of Sydney School of Physics Foundation's Grand Challenge initiative.
S.M. acknowledges support from the Australian Research Council (ARC) via the Future Fellowship, `Emergent many-body phenomena in engineered quantum optical systems', project no. FT200100844.
B.D.F. acknowledges support from the Australian Research Council (FT240100418).
\end{acknowledgments}

\appendix

\section{Data pre-processing details}
\label{appendix:data-pre-proc-details}

Here we discuss the details of the data transformations we applied to each raw time-series dataset prior to encoding them with the feature map in Eq.~\eqref{eq:basis-fns}. 
For a given time-series dataset of $N$ instances and $T$ data points (samples) per instance, represented by an $N \times T$ data matrix $\boldsymbol{X}$, we used an outlier-robust sigmoid transformation \cite{Fulcher2013:HighlyComparativeTimeseriesa}:
\begin{equation} \label{eq:sigmoid}
    \boldsymbol{X'} = \left(1 + \exp{-\frac{\boldsymbol{X} - m_{\boldsymbol{X}}}{r_{\boldsymbol{X}} / 1.35}}\right)^{-1},
\end{equation}
where $\boldsymbol{X'}$ is the normalized time-series data matrix, $\boldsymbol{X}$ is the un-normalized time-series data matrix, $m_{\boldsymbol{X}}$ is the median of $\boldsymbol{X}$ and $r_{\boldsymbol{X}}$ is its interquartile range (IQR).
In the case of a scaled robust sigmoid transform, an additional `min--max' linear transformation is applied to $\boldsymbol{X'}$ such to rescale it to the target range $[a, b]$:
\begin{equation}
    \boldsymbol{X''} = \left(b - a \right)  \cdot \frac{\boldsymbol{X'} - x'_{\text{min}}}{x'_{\text{max}} - x'_{\text{min}}} + a,
    \label{eq:minmax}
\end{equation}
where $\boldsymbol{X''}$ is the scaled robust-sigmoid transformed data matrix, $x'_\text{min}$ and $x'_\text{max}$ are the minimum and maximum of $\boldsymbol{X'}$, respectively. 
As discussed in Sec.~\ref{sec:encoding-methodology} for time-series imputation we only used the 
linear transformation in Eq.~\ref{eq:minmax}, but applied to the raw data matrix $\boldsymbol{X}$ rather than the sigmoid-transformed data matrix $\boldsymbol{X'}$. 
For classification, we used the scaled robust sigmoid transformation as defined in Eq.~\eqref{eq:minmax}.

When we evaluated a trained model on unseen data, the unseen data was normalized using the values $m_{\boldsymbol{X}}$, $r_{\boldsymbol{X}}$, $x'_{\text{min}}$, and $x'_{\text{max}}$ estimated from the training dataset to avoid information leakage between the training and test sets.
An undesirable implication of this necessary processing is that occasionally some test time series were mapped outside of the range $[a, b]$.
For each of these time-series instances:
\\
\begin{enumerate}[label=\roman*), nosep]
    \item If the minimum was less than $a$, the time series was shifted up so that it had a minimum value of $a$. 
    \item After shifting the time series, if the new maximum was larger than $b$, the time series was linearly rescaled so that the maximum was at most $b$.
\end{enumerate}
\vspace{5pt}
These outliers had no meaningful bearing on classification performance. For example, for the default UCR train/test split of the ECG dataset, only $5$ out of the $100$ sigmoid-transformed time series in the test set were shifted up by $0.1\%$, $0.1\%$, $0.2\%$, $1\%$, and $1.7\%$, respectively. 
No time series were rescaled due to maximums exceeding the encoding domain after shifting.
For the default UCR train/test split of the Power Demand dataset, only $34$ out of the $1029$ sigmoid-transformed test time series were altered.
Specifically, $13$ time-series instances were shifted up, the largest of which were $1.9\%$, $2.1\%$, $2.1\%$, $4.4\%$ and $5.1\%$. 
A total of $23$ time series were rescaled, with the five largest rescales being by $2.6\%$, $3.0\%$, $4.1\%$, $4.5\%$ and $8.8\%$. 

When evaluating imputation performance in Sec.~\ref{sec:imputation-experiments}, the mean absolute error (MAE) defined in Eq.~\eqref{eq:MAE} was computed on the raw (non-processed) data to ensure errors were presented in meaningful and comparable units across all imputation algorithms.

\section{Impact of artificial broadening from finite basis representation on conditional imputation}
\label{appendix:enc-error}
In this work, we train an MPS to encode a probability distribution of time-series amplitudes.
As these amplitudes are real-valued and continuous, with each $x_t \in \mathbb{R}$, they must first be mapped to discrete quantum states, $\ket{x_t}$, to be compatible with the MPS framework.
To achieve this discretization, we project the amplitudes onto a truncated basis set of $d$ functions, $\phi_t(x_t) = \{b_{1}(x_t), b_{2}(x_t),\ldots, b_{d}(x_t) \}_t$.
Here, the choice of $d$ determines the resolution of the discretization, with larger $d$ enabling a finer approximation of the original continuous time-series amplitudes.

\begin{figure}[t]
    \centering
    \includegraphics[width=1.0\linewidth]{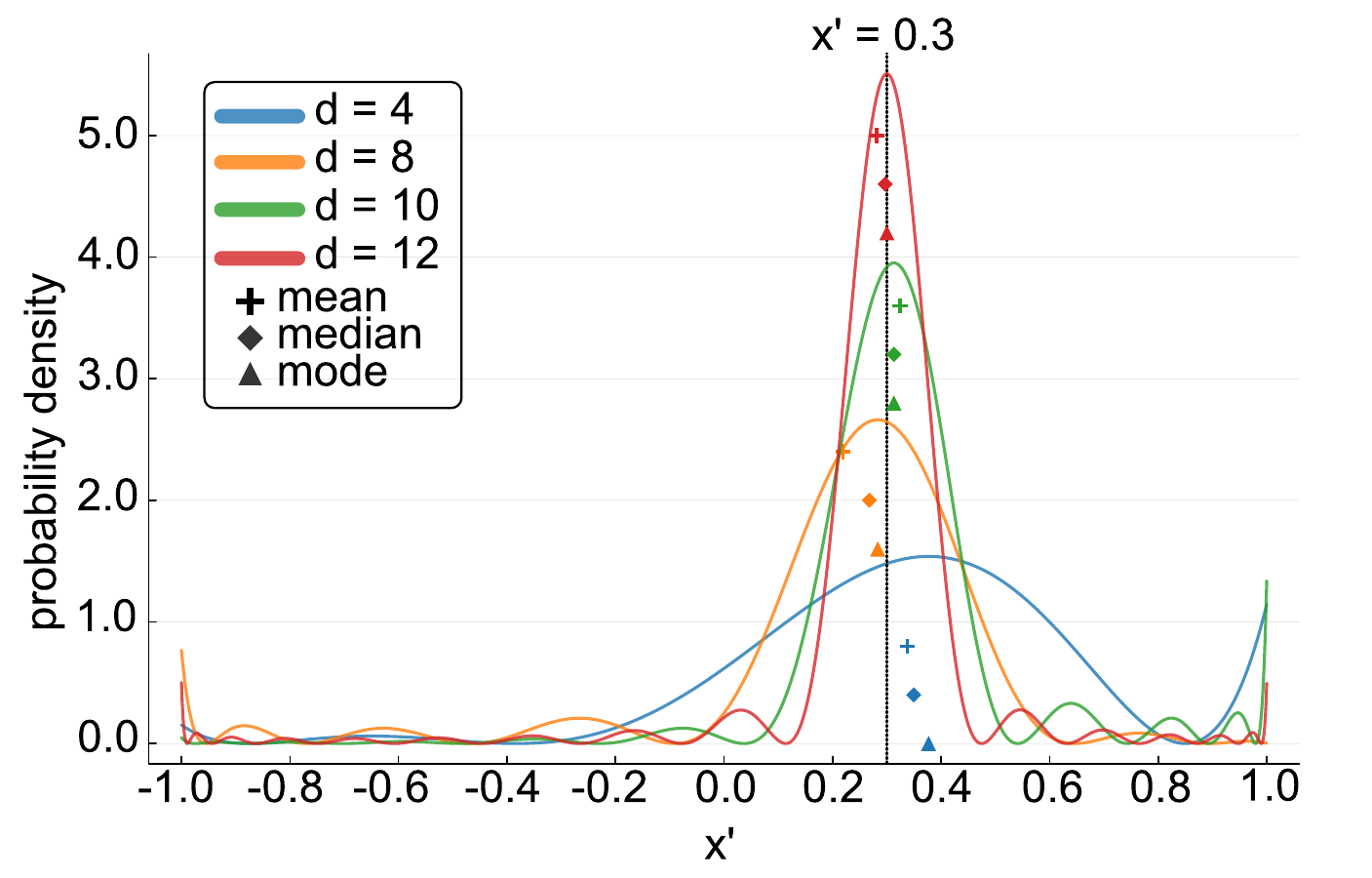}
    \caption{\textbf{Distribution broadening due to the finite basis representation of a continuous value.}
    Given a continuous value in the compact domain $x' \in [-1, 1]$, e.g., $x' = 0.3$, represented by the vertical black line, the corresponding conditional distribution under the finite Legendre basis representation is shown for varying MPS physical dimension $d$.
    Three measures of the central tendency of a distribution: mean (cross marker), median (diamond marker), and mode (triangle marker) are shown for $d = 4$ (blue), $d = 8$ (gold), $d = 10$ (green), and $d = 12$ (red).}
    \label{fig:dist-broadening}
\end{figure}

As a result of this discretization, the reduced density matrix (RDM) at each site can assign artificially inflated probabilities to states that would otherwise be unlikely under the true data distribution.
This effect is depicted visually in Fig.~\ref{fig:dist-broadening}, which shows the `artificial broadening' of the conditional distribution, given a ground-truth continuous value, $x' = 0.3$ (vertical black line), encoded as a discrete state via a Legendre feature map with $d$ basis functions, as in Eq.~\eqref{eq:basis-fns}.
For low physical dimension $d = 4$ (blue), corresponding to a coarser approximation, the distribution is broad and skewed.
Repeatedly sampling from this distribution will tend to yield estimates far from $x'=0.3$ (vertical black line).
As $d$ increases (e.g., $d = 12$, red) -- yielding a more fine-grained approximation of the continuous value -- the mass concentrates around $x' = 0.3$, and measures of central tendency such as the median (red diamond marker) and mode (red triangle marker) converge toward the true value, with the latter two summary statistics exhibiting the greatest robustness.

\begin{figure*}[htbp!]
    \centering
    \includegraphics[width=1.0\linewidth]{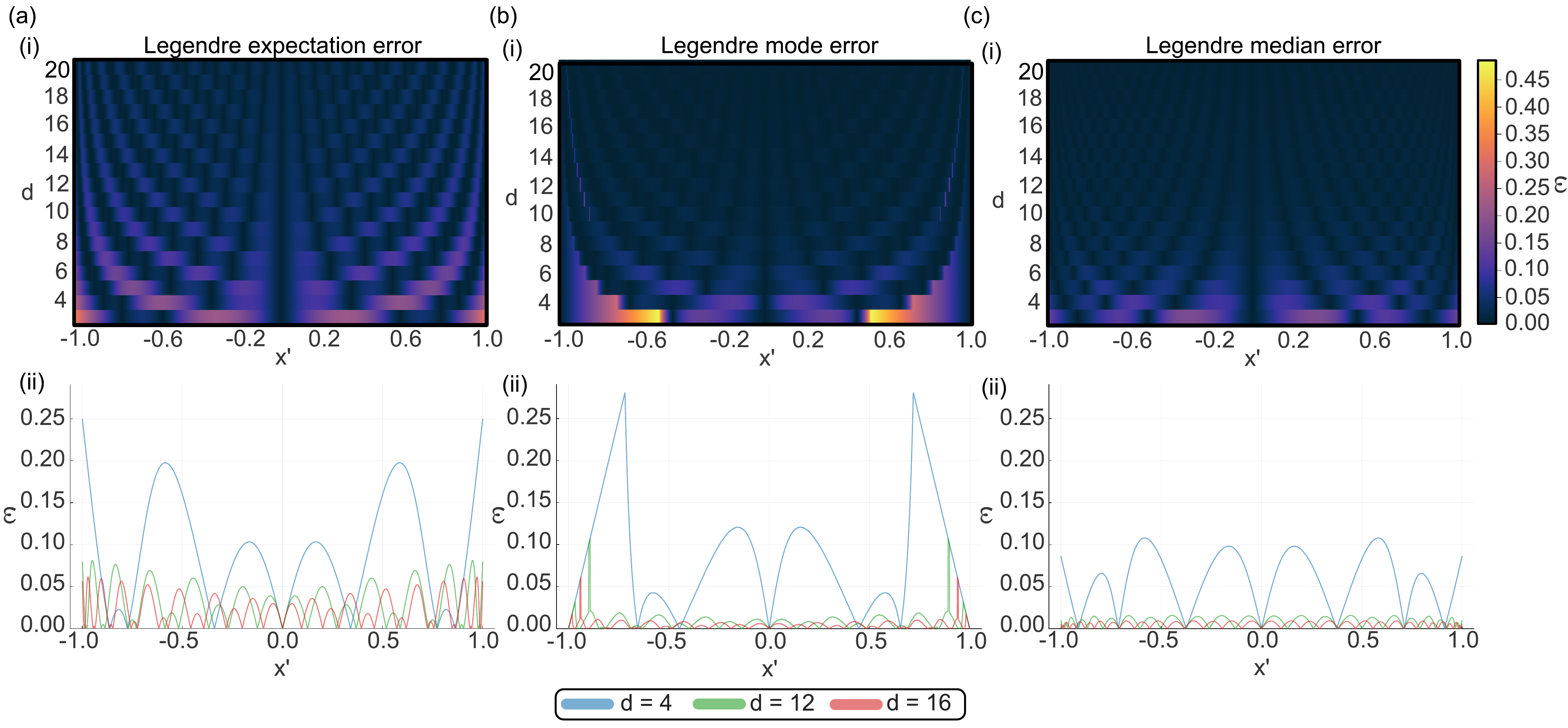}
    \caption{\textbf{Encoding error of continuous values over the Legendre domain under a finite basis approximation}.
    The absolute error between a known value $x'$ in the encoding domain, and a summary statistic for the probability distribution corresponding to the conditional density matrix $ \rho = \phi^\dagger (x')\phi(x')$ (note that the Legendre feature map $\phi$ is time independent), which we refer to as the encoding error $\epsilon_d$ (as in Eq.~\eqref{eq:enc-error}) is shown for \textbf{(a)} the mean (expectation); \textbf{(b)} the mode; and, \textbf{(c)} the median.
    \textbf{Top row (i):} shows the encoding error, indicated by the color scale, when using a measure of central tendency as a single point estimator
    for varying $d$ and $x'$ across the compact Legendre encoding domain $x' \in [-1, 1]$.
    \textbf{Bottom row (ii):} each column -- (a)(ii), (b)(ii), (c)(ii) shows the encoding error for selected values of physical dimension, $d = 4$ (blue), $d=12$ (green) and $d=16$ (red), corresponding to the expectation error in (a)(i), mode error in (b)(i), and median error in (c)(i), respectively.}
    \label{fig:enc-error}
\end{figure*}

In this work, the imputation problem we address involves inferring a single point estimate for each missing time-series sample.
In this setting, rather than sampling from the conditional distribution, for a sufficiently large $d$, we aimed to use a measure of central tendency as our point estimate.
We then defined an `encoding error' $\epsilon_d$ to quantify the discrepancy introduced by the finite basis approximation:
\begin{equation} \label{eq:enc-error}
    \epsilon_d = \lvert x_i - \hat{x}_{i, d} \rvert\,,
\end{equation} 
where $x_i$ is a continuous value in the encoding domain, and $\hat{x}_{i, d}$ is a summary statistic capturing the central tendency of the conditional probability distribution at site $i$ after encoding $x_i$ via the feature map $\phi_i(x_i)$ with $d$ basis functions.
In Fig.~\ref{fig:enc-error}, we compare the encoding error for three measures of central tendency -- the expectation (mean), median, and mode -- as a function of $d$ and uniformly spaced values in the Legendre encoding domain ($x_i \in [-1, 1]$).
As expected, the mean, shown in Fig.~\ref{fig:enc-error}(a), gives rise to large encoding errors across the domain, particularly for smaller $d$, as a result of its sensitivity to the artificial broadening of the distribution.
The mode (in Fig.~\ref{fig:enc-error}(b)), while competitive for higher $d$, exhibits significant error spikes near the domain boundaries which can lead to erroneous imputations.
The median, shown in  Fig.~\ref{fig:enc-error}(c), was found to consistently outperform other summary statistics, providing the lowest average error across the encoding domain.
Motivated by these empirical findings, we chose to adopt the median for MPS-based time-series imputation, given that it best mitigates the effects of artificial broadening and ensures a sufficiently representative single point estimate from the encoded distribution.

\section{MPSTime hyperparameter tuning}
\subsection{Imputation hyperparameter tuning}
\label{appendix:imputation-hyperparm-opt}
For our imputation experiments in Sec.~\ref{sec:imputation-experiments}, the MPS physical dimension $d$, maximum bond dimension $\chi_{\rm max}$, and learning rate $\eta$ were determined by hyperparamater optimization.
For the learning rate $\eta$, we searched in logarithmic space to efficiently cover multiple orders of magnitude.
We note that in Sec.~\ref{sec:imputation-synthetic} $\chi_{\rm max}$ was held fixed, as we sought to investigate the relationship between bond dimension (model expressibility) and imputation accuracy.
The number of sweeps was fixed to $n_{\rm sweeps} = 10$ across all experiments, and we used a min--max pre-processing transformation for the data, as described in Sec.~\ref{sec:encoding-methodology}.
To search the space of hyperparameters bounded by the ranges defined in Table.~\ref{tab:mps-imputation-parameter-ranges} we used an off-the-shelf implementation of Latin Hypercube Sampling (LHS) \cite{Bates2004:FormulationOptimalLatin} with 5-fold cross-validation (CV).
LHS is a random search variant suited to efficiently sampling from multi-dimensional distributions.
For each CV fold, $250$ MPS models (i.e., unique hyperparameter combinations) were evaluated based on their mean MAE across the 5\% to 95\% data missing range, averaged across all held-out validation instances.

\begin{table}[H]
    \caption{\textbf{MPSTime imputation hyperparameter ranges for each dataset in Sec.~\ref{sec:imputation-experiments}.}}
    \centering
    \begin{tabular}{ c  c  c  c  c  c}
         Hyperparameter & \textbf{NTS1-4} & \textbf{NTS5} & \textbf{\textsuperscript{\S}Empirical}\\
         \hline
         $d$ & [5, 15] & [5, 15] & [5, 15] \\
         $\eta$ & [0.001, 0.5] & [0.001, 0.5] & [0.001, 0.5] \\
         $\chi_{\rm max}$ & [20, 80]\textsuperscript{*} & [20, 160] & [20, 40] \\
    \end{tabular}
    \label{tab:mps-imputation-parameter-ranges}
    \footnotesize NTS1--4 refers to datasets NTS1, NTS2, NTS3, NTS4 described in Sec.~\ref{sec:dataset-details-synthetic}.
    \textsuperscript{*}For NTS1, we hold $\chi_{\rm max}$ fixed at $20, 30, 40$.
    \textsuperscript{\S}Empirical refers to the shared hyperparameter ranges for the three real-world datasets: ECG, Power Demand, and Astronomy, described in Sec.~\ref{sec:real-world-dataset-details}. 
\end{table}

\subsection{Classification hyperparameter tuning}
\label{appendix:hyperparam-tuning}
For classification, the physical dimension $d$, maximum bond dimension $\chi_{\rm max}$ and learning rate $\eta$ were determined using the same hyperparameter tuning search algorithm in Appendix~\ref{appendix:imputation-hyperparm-opt} with 5-fold CV.
The number of sweeps was fixed to $n_{\rm sweeps} = 10$ across all experiments, and we used a scaled outlier robust sigmoid transformation to normalize the raw time-series data, as described in Sec.~\ref{sec:encoding-methodology}.
We evaluated $250$ MPS models for misclassification rate (i.e., 1 - accuracy) on each CV fold.
The particular hyperparameter ranges we searched for each empirical dataset are summarized in Table~\ref{tab:hyperparams-classification}.
\begin{table}[H]
    \caption{\textbf{MPSTime imputation hyperparameter ranges for each dataset in Sec.~\ref{section:classification-results}.}}
    \centering
    \begin{tabular}{ c  c  c  c }
         Hyperparameter & \textbf{ECG} & \textbf{Power Demand} & \textbf{Astronomy} \\
         \hline
         $d$ & [2, 15] & [2, 15] & [2, 15] \\
         $\eta$ & [0.001, 0.5]\textsuperscript{\S} & [0.001, 0.5]\textsuperscript{\S} & [0.001, 0.5]\textsuperscript{\S} \\
         $\chi_{\rm max}$ & [20, 40] & [20, 40] & [20, 40] \\
    \end{tabular}
    \footnotesize \textsuperscript{\S}$\eta$ was searched in logarithmic space to efficiently cover multiple orders of magnitude.
    \label{tab:hyperparams-classification}
\end{table}

\section{Imputation algorithm details}
\label{app:interp_alg}
\begin{figure}[h]
    \centering
    \includegraphics[width=1.0\linewidth]{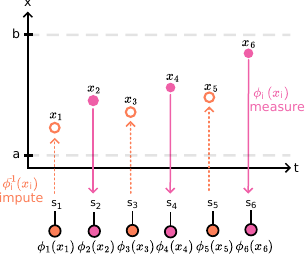}
    \caption{\textbf{A toy imputation problem involving six time-series samples}.
    We consider three observed data points (solid pink markers)
    $x_2$, $x_4$, $x_6$, and three unobserved (missing) data points, $x_1$, $x_3$, $x_5$ (open orange markers).
    Observed values in the encoding domain $x_t \in [a, b]$ are mapped to discrete quantum states (pink vectors) via the encoding $\phi_i(x_i)$ i.e., $s_i = \phi_i(x_i)$.
    The MPS is then conditioned on the observed values by making a series of projective measurements at the sites corresponding to the observed states, $s_2$, $s_4$, $s_6$.
    The unobserved states (orange vectors) are determined from the conditional probability distribution encoded by the partially conditioned MPS, and transformed back to time-series values in the encoding domain (orange markers) by an inverse transformation, $\phi_{i}^{-1}$.}
    \label{fig:toy-interp-schematic}
\end{figure}

Here we provide details on the specific steps of the MPSTime imputation algorithm summarized in Sec.~\ref{sec:imputation} of the main text. 
To elucidate the key steps of the proposed algorithm, we consider a 6-site MPS trained on time-series instances of length $T = 6$ samples from some class $C$, shown in Fig.~\ref{fig:toy-interp-schematic}.
Given an unseen time-series instance of data class C that is incomplete, for example, only the values $x_2$, $x_4$, $x_6$ are known, the goal of imputation is to estimate the unknown values (i.e., to `fill in the gaps') - in this case, $x_1$, $x_3$, $x_5$.
Our approach is performed in two main steps:
(i) the trained MPS is conditioned on known values by first transforming them into quantum states and then projecting the MPS onto these known states;
(ii) using the conditioned MPS, the unknown values are then determined by sequentially computing and sampling from a series of single-site reduced density matrices.

\subsection{Projecting the trained MPS onto known states}
\label{appendix:imputation-projection}
The first step of the algorithm is to project the MPS -- through a sequence of iterative site-wise measurements -- into a subspace which can then be used to impute missing data.
We assume the MPS $W$, which is trained to approximate the underlying joint PDF of class C time-series data, is L2 normalized. 
This normalization condition can be shown diagrammatically in Penrose graphical notation \cite{penrose1971applications}) as:
\begin{figure}[H]
    \centering
\includegraphics[width=0.85\linewidth]{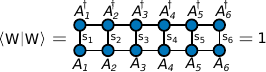}
\end{figure}
\noindent where $s_1, s_2,\dots,s_6$ correspond to the physical indices at each site.
Using the feature-mapped state associated with the first known value of the time series, $\phi_2(x_2)$, a projective measurement is made at the corresponding MPS site $A_2$ by contracting its physical index with that of the state vector $\phi_2(x_2)$ (shown in pink):
\begin{figure}
    \centering
\includegraphics[width=0.89\linewidth]{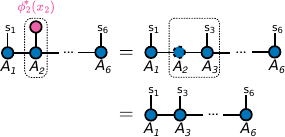}
\end{figure}
To ensure the updated MPS continues to represent a valid conditional probability distribution, the entries in the tensor corresponding to the measured site (dashed circle) are divided through by the factor $\sqrt{P(s_2)}$, yielding a normalized MPS.
To compute the marginal probability $P(s_2)$, we perform a contraction operation with two copies of the MPS (one conjugated), first tracing over all remaining sites, and then contracting the state vector of interest $\phi_2(x_2)$ with its corresponding MPS site (on both copies), shown diagrammatically as:
\begin{figure}[H]
    \centering
    \includegraphics[width=0.68\linewidth]{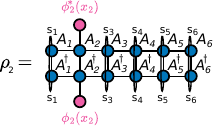}
\end{figure}
\noindent Note that in the main text, we refer to $P(x_i) = P(\phi_i(x_i))$ in order to reduce notational clutter.
The rescaled tensor is then contracted over the bond index with the adjacent site to yield an updated MPS that is one site shorter.
Having projected the MPS onto the known state at the first site, the updated MPS now approximates the PDF conditioned on the known time-series value at $t = 2$: 
\begin{figure}[H]
    \centering
\includegraphics[width=0.7\linewidth]{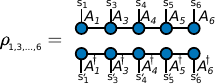}
\end{figure}
\noindent Here, we use open physical indices to represent the entire structure of the joint probability distribution over all possible configurations that $s_1, s_3, \dots, s_6$ can take.
Using the known value of the time series at $t = 4$ (pink vector), one then proceeds to make a second projective measurement of the MPS onto its corresponding state at the fourth site $A_4$ (now the third site), before normalizing the updated tensor by dividing through by $\sqrt{P(x_4)}$ and contracting with its neighboring site:
\vspace{-8pt}
\begin{figure}[H]
    \centering
    \includegraphics[width=0.85\linewidth]{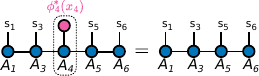}
\end{figure}
\vspace{-8pt}
Here, we show the resulting 4-site MPS which approximates the PDF conditional on having measured the states $s_2, s_4$.

This process of making projective measurements, normalizing, and contracting with the neighboring site proceeds recursively for all remaining known data points, yielding a final MPS which approximates the PDF conditioned on all known time-series values:
\vspace{-8pt}
\begin{figure}[H]
    \centering
    \includegraphics[width=0.48\linewidth]{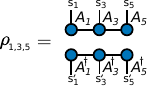}
\end{figure}
\vspace{-8pt}
\noindent with sites corresponding to the unknown states (time-series data points) that are to be imputed. 

\subsection{Imputing missing time-series data}
The second step of the MPSTime imputation algorithm involves determining the values of the unobserved data points by estimating their respective values from the conditional PDF represented by the MPS. 
Here we use a sequential sampling approach which involves computing $d \times d$ single-site reduced density matrices.

We obtain the reduced density matrix $\rho_1$ of the first imputation site $A_1$ by tracing out the remaining sites $A_3$, $A_6$, shown diagrammatically as:
\vspace{-8pt}
\begin{figure}[H]
    \centering
    \includegraphics[width=0.5\linewidth]{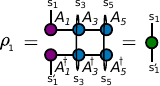}
\end{figure}
\vspace{-8pt}
The probability density function (pdf) of the continuous variable $X$ is evaluated using the reduced density matrix $\rho$ as follows:
\begin{equation}
    \label{eq:pdf}
    \textrm{pdf}_{X}(x) = \phi^\dagger(x)\rho \, \phi(x).
\end{equation}
To sample a value for $x_1$ from the continuous distribution with this pdf, i.e., $x_1 \sim \textrm{pdf}_X(x)$, one would typically resort to inverse transform sampling (see \cite{Mossi2024:MatrixProductState} and \cite{ Meiburg2023:GenerativeLearningContinuous} for more details).
However, for the purposes of imputation, our primary focus is on obtaining the best single-point estimate and its associated uncertainty, given the known data.
This allows us to work directly with the reduced density matrix $\rho$, circumventing the need to generate individual samples. 
Using $\rho_1$, we calculate the median of the continuous random variable $X$, which serves as our best point estimate for $x_1$: 
\begin{equation}
    \text{median}(X) = \hat{x} \quad \text{such that} \quad \frac{1}{Z}\int_{-\infty}^{\hat{x}} f_X(x) \, dx = 0.5\,,
\end{equation}
where $\textrm{pdf}_X(x)$ is defined in Eq.~\ref{eq:pdf} and $Z$ is a normalization factor:
\begin{equation}
    Z = \int_{a}^{b} \textrm{pdf}_X(x)dx\,,
\end{equation}
which ensures the pdf remains normalized over the support of the encoding domain $[a, b]$ ($[-1, 1]$ for the Legendre basis).
For details about our choice of the median as the best single point estimate, see Appendix.~\ref{appendix:enc-error}.

To quantify the uncertainty in our point estimate $\hat{x}$ we use the weighted median absolute deviation (WMAD):
\begin{equation}
    \label{eq:wmad}
    \text{WMAD}(\hat{x}) = \text{median}(\lvert x_i - \hat{x}\rvert, w_i)\,,
\end{equation}
where $\hat{x}$ is the median (our point estimate for $x_1$), $x_i$ are discretized values of the random variable $X$ within the encoding domain $[a,b]$, and the weights $w_i$ are proportional to the probability densities $w_i \propto \textrm{pdf}_X(x_i)$.

Using the value of the point estimate $x_1$ and the associated state $\phi_1(x_1)$, a projective measurement is made at the corresponding MPS site $A_1$, completing one step of the imputation algorithm.
The process then continues recursively -- at each step, obtaining the single site reduced density matrix $\rho_1$, computing the point estimate $x^*_1$, and performing a projective measurement -- until all remaining unknown values have been determined:
\vspace{-8pt}
\begin{figure}[H]
    \centering
    \includegraphics[width=0.73\linewidth]{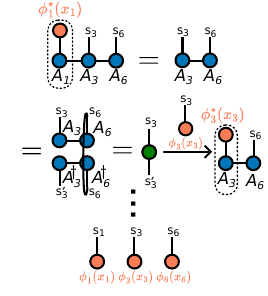}
\end{figure}
\vspace{-8pt}
\noindent To transform the imputed states (orange vectors) back to their rescaled time-series values, one can apply the inverse of the feature map $\phi_t^{-1}(x)$. 
An additional transformation can be applied to convert the rescaled time-series (in the encoding domain) back to their respective amplitudes in the original data domain.

\section{Inverse transform sampling}
\label{appendix:inverse-transform-sampling}
To generate time-series trajectories, as in Fig.~\ref{fig:mps_vs_data_dist}, we used an inverse transform sampling (ITS) approach augmented with a rejection scheme to mitigate the effects of the encoding error discussed in Sec.~\ref{appendix:enc-error}.
Specifically, due to the finite basis approximation of (original continuous) time-series values, we observe an error which manifests as an artificial broadening of the conditional distribution of time-series values at each MPS site.
As a consequence of this broadening, the conditional distribution can assign non-negligible probabilities to time-series values that would otherwise be highly improbable, as depicted in Fig.~\ref{fig:dist-broadening}.
When generating trajectories through conditional site-wise sampling -- starting at the first site and proceeding sequentially to the end -- these errors can propagate and compound, leading to trajectories that are not representative of the underlying joint distribution.
While increasing $d$, the number of basis functions in the encoding (as in Eq.~\eqref{eq:basis-fns}), can mitigate this error in principle, achieving meaningful reductions in distribution broadening would require values that render ITS computationally intractable for most applications.
Instead, we use a rejection scheme to discard highly improbable values based on their deviation from the median (i.e., weighted median absolute deviation in Eq.~\eqref{eq:wmad}) of the conditional distribution.
Our choice of the median was motivated empirically from the analysis presented in Appendix~\ref{appendix:enc-error}.

To perform inverse transform sampling (ITS), given an MPS site $A_i$ corresponding to an unobserved time-series value $x_i$, we use the following procedure to sample the value from the corresponding conditional distribution, i.e., $x_i \sim pdf_i(x)$.
First, we evaluate the cumulative distribution function, $F_i(x)$, as:
\begin{equation} \label{eq:rs-cdf}
     F_i(x) = \frac{1}{Z} \int_{a}^{x} \phi^\dagger(x') \rho_i \phi_i(x') \ud x'\,,
\end{equation}
with $Z$ chosen so that $F_{s_i}(b) = 1$, and where $a$ is the lower bound on the support of the encoding domain $[a, b]$. 
Next, we sample a random value from a uniform distribution defined on the interval $[0, 1]$:
\begin{equation}
    u \sim U(0, 1)\,.
\end{equation}
Using the inverse cumulative distribution function, $F_i^{-1}(u)$, we select the value $x_i$ such that $F_i(x_i) = u$.
We then apply the WMAD-based rejection criteria described in Algorithm~\ref{alg:its}.

The selection of the rejection threshold factor $\alpha$ requires careful consideration.
A threshold that is overly restrictive (i.e., small $\alpha$) would constrain the sampling space, potentially introducing bias by suppressing variability that is a feature of the true encoded distribution.
Conversely, a less-stringent threshold (i.e., large $\alpha$) would fail to adequately address the artificial broadening effects.
Through empirical investigation, we identified $\alpha = 2.0$ (i.e., $2 \cdot \rm WMAD)$ as an effective heuristic, which provides a balance between the two aforementioned extremes.

\begin{algorithm}[H]
\caption{Single trajectory time-series generation.}
\label{alg:its}
\begin{algorithmic}[1]
\Statex
    \State Compute the conditional PDF, $\textrm{pdf}_i(x)$, at MPS site $A_i$.
    \State Compute the conditional CDF as in Eq.~\eqref{eq:rs-cdf}.
    \State Generate a uniform random number $u \sim U(0,1)$.
    \State Find $x_i = F_i^{-1}(u)$ such that $F_i(x_i) = u$.
    \State Compute the median $m_i$ and WMAD $wm_i$ of the conditional distribution $pdf_i(x)$.
    \State Evaluate the deviation of $x_i$ from $m_i$:
            \[
           \Delta_i = \lvert x_i - m_i\rvert\,.
           \]
    \While{$\Delta_i > \alpha \cdot wm_i$}
    \Comment{Rejection step}
        \State Generate a new $u \sim U(0,1)$.
        \State Find $x_i = F_i^{-1}(u)$.
        \State Evaluate $\Delta_i$.
    \EndWhile
    
    \State Accept $x_i$ as the $i$-th value in the trajectory.
\end{algorithmic}
\end{algorithm}
\section{Real-world time-series datasets}
\label{appendix:dataset-details}
\begin{figure}[H]
    \centering
    \includegraphics[width=1.0\linewidth]{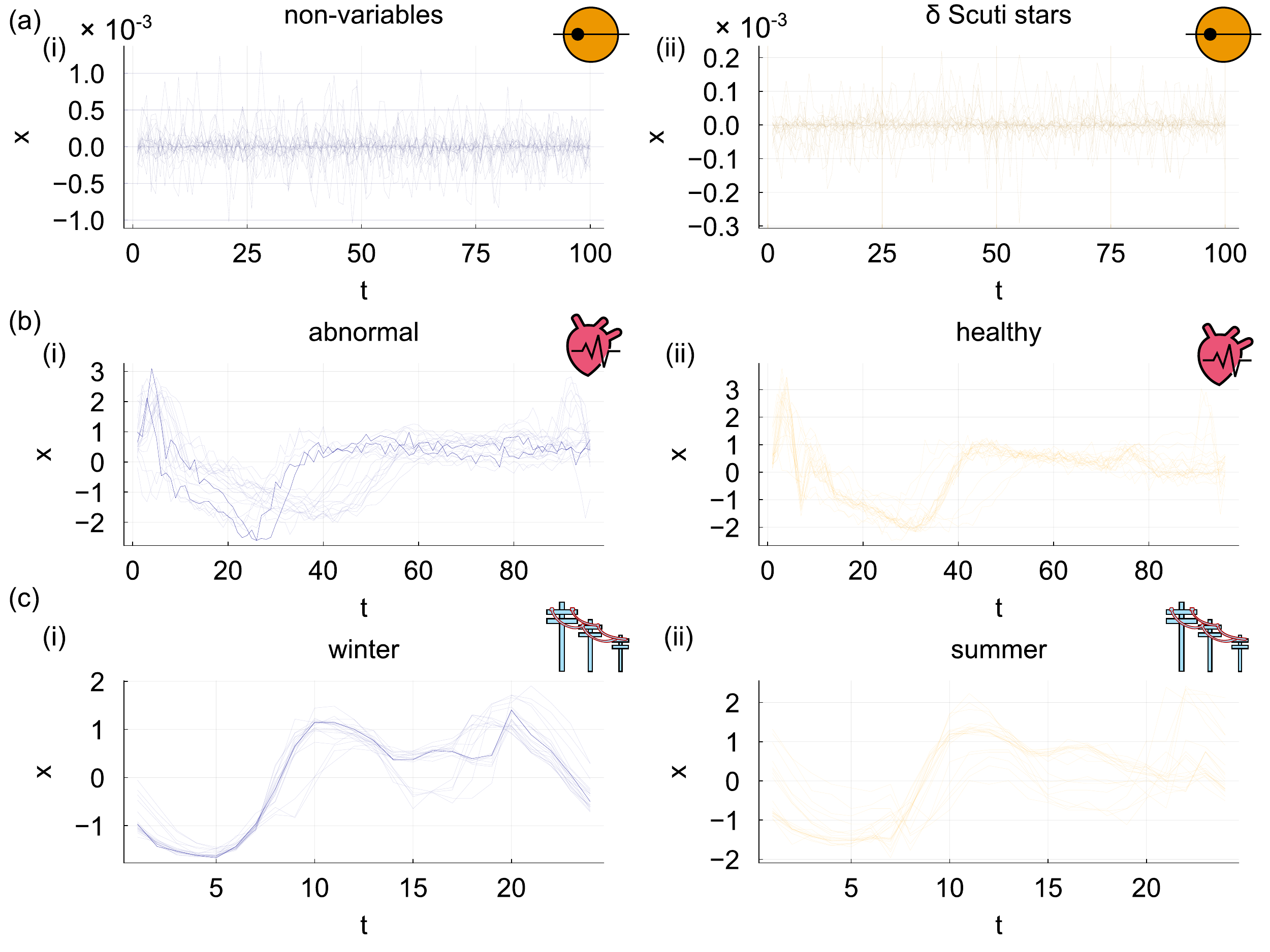}
    \caption{\textbf{Time-series instances for each empirical dataset in the classification task.}
    Time-series instances from each class are plotted on the same axis (and in the original data domain) for the three datasets: \textbf{(a)} Astronomy (2-class subset of \texttt{KeplerLightCurves}); 
    \textbf{(b)} ECG (\texttt{ECG200});
    and \textbf{(c)} Power Demand (\texttt{ItalyPowerDemand}).
    In each panel, we plot all instances from (i) class 0 (blue time-series instances), and (ii) class 1 (gold time-series instances), for each binary classification problem.}
    \label{fig:example-ts}
\end{figure}

Here we provide further details on the three University of California, Riverside (UCR) Time Series Archive \cite{Dau2019:UCRTimeSeries} datasets used for the classification task in Sec.~\ref{section:classification-results}, each representing key domain of application: (i) ECG (medicine); (ii) Power Demand (industry); and (iii) Astronomy (physics). 
Representative time-series examples from each dataset, prior to rescaling, are shown in Fig.~\ref{fig:example-ts}.

\begin{figure*}
    \centering
    \includegraphics[width=1.0\linewidth]{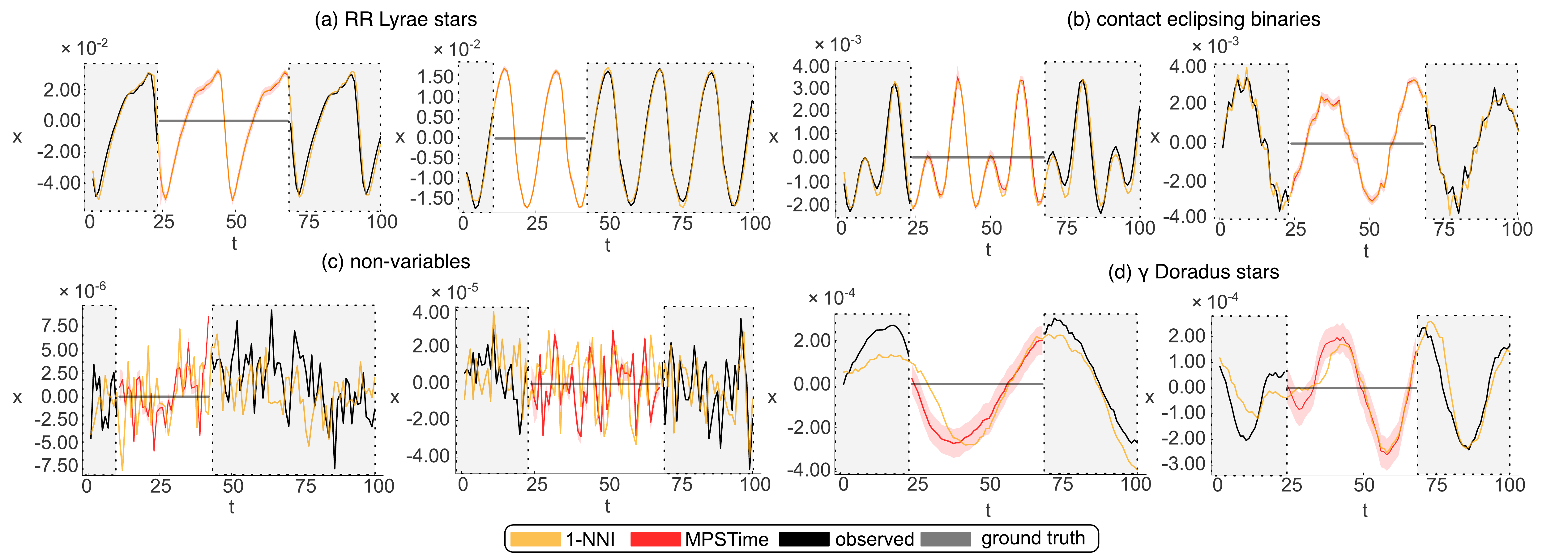}
    \caption{\textbf{MPSTime imputations on unseen Kepler light curve time series with real data gaps arising from satellite transmission interruptions.}
    The panels depict examples of MPSTime ($\chi_{\rm max} = 35$, $d = 12$) imputations (red) and a 1-NNI benchmark (gold) on held-out instances with real data gaps from four classes of star system: \textbf{(a)} RR Lyrae stars, \textbf{(b)} contact eclipsing binaries, \textbf{(c)} non-variables, and \textbf{(d)} $\gamma$ Doradus stars.
    The black points correspond to observed time-series values, while the gray horizontal lines correspond to segments of missing data, which were imputed in the original dataset \cite{Barbara2022:ClassifyingKeplerLight} using the mean of the observed values.
    }
    \label{fig:kepler_no_gt}
\end{figure*}

\subsection{Imputation on real-world data gaps}
For our investigation of the Astronomy dataset in Sec.~\ref{sec:imputation-experiments} of the main text, we excluded time-series windows containing real missing values as there exists no ground-truth for evaluation.
Nevertheless, such examples provide a compelling demonstration of the real-world importance of imputation algorithms for cases when there are durations over which sensors are not recording data.
For completeness, here we showcase selected examples of MPSTime imputations on real data gaps from four classes of star system (comprising the original KeplerLightCurves dataset): (i) RR Lyrae stars; (ii) contact eclipsing binaries; (iii) non-variables; and (iv) $\gamma$ Doradus stars.
We trained MPSTime on length $T = 100$ sample non-overlapping windows of a single time-series instance, as in Sec.~\ref{sec:real-world-imputation-results}.
Each example, shown in Fig.~\ref{fig:kepler_no_gt}, compares the MPSTime imputation (red) with a 1-Nearest Neighbor Imputation (1-NNI) (gold) benchmark.
Notably, while MPSTime and 1-NNI produce similar predictions for uniformly repetitive periodic light curves, such as RR Lyrae stars and contact eclipsing binaries (Figs~\ref{fig:kepler_no_gt}a and b, respectively), the predictions of each algorithm diverge for less uniform patterns, such as non variable and $\gamma$ Doradus stars (Figs~\ref{fig:kepler_no_gt}c and d, respectively).

\section{Time-series classification benchmarks}
\label{appendix:tsc-baselines}
For classification performance we selected three benchmarks, each representing a key category of time-series classification algorithm: (i) distance-based (1-NN-DTW); (ii) deep learning (InceptionTime); and (iii) a hybrid ensemble (HIVE-COTE v2.0).
Specifically:
\begin{itemize}[noitemsep]
    \item \textbf{1-NN-DTW}: Nearest neighbor using a Dynamic Time Warping (DTW) similarity measure. 
    This is the most commonly used standard classical benchmark in much of the time-series classification literature \cite{Bagnall2017:GreatTimeSeries}.
    \item \textbf{InceptionTime}~\cite{Fawaz2020:InceptionTimeFindingAlexNet}: A state-of-the-art deep learning classifier consisting of an ensemble of five Convolutional Neural Network (CNN) models.
    \item \textbf{HC2}~\cite{Middlehurst2021:HIVECOTENewMeta}: HC2 (HIVE-COTE V2.0) is an ensemble of classifiers, each built on different time-series data representations, including phase-independent shapelets, bag-of-words based dictionaries, among others.
    At the time of the analysis, HC2 is state-of-the-art for classification accuracy on the UCR time-series repository \cite{Middlehurst2024:BakeReduxReview}.
\end{itemize}
For the Astronomy dataset, we used off-the-shelf implementations of the 1-NN-DTW, InceptionTime, and HC2 classifiers that are accessible at the Python \texttt{aeon} (\texttt{v0.11.1}) library \cite{Middlehurst2024:AeonPythonToolkit}.
Time-series instances were $z$-score normalized (each instance along its time axis), as recommended for univariate time-series classification \cite{Dau2019:UCRTimeSeries}.
For the InceptionTime benchmarks, we used the hyperparameter values from the original implementation \cite{Fawaz2020:InceptionTimeFindingAlexNet}, which are set as defaults in \texttt{aeon}, and trained the model for 1500 epochs on an NVIDIA L40S GPU.
To facilitate a fair comparison, we allowed HC2 the same computational resources as InceptionTime by limiting the maximum training time to match the time taken to train InceptionTime on the same hardware.
We also used the default hyperparameters for HC2 as provided in the \texttt{aeon} implementation.

\section{Time-series classification training times}
\label{appendix:tsc-training-times-comparison}
To compare training-time performance, we recorded the wall-clock time required to train each algorithm on the ECG dataset using representative hyperparameters from the classification benchmark in Sec.~\ref{section:classification-results}.
Here, we refer to ``training time" as the time required to execute each classifier's \texttt{fit} call on the training data.
HIVE-COTE V2.0 (HC2) was excluded from the analysis since due to its algorithmic formulation, computation time results were not fairly comparable.
To ensure our measurements only reflected steady-state performance, we perform one initial run for each algorithm and discard it as a burn-in. 
Table~\ref{tab:ecg-fit-time} reports the mean and standard deviation of training times over $30$ train--test resample folds.
On this dataset MPSTime required on average $1.90 \times 10^{1}$\,s to train -- nearly two orders of magnitude faster than InceptionTime ($1.72 \times 10^{3}$\,s), despite the latter leveraging GPU acceleration on specialized hardware (NVIDIA L40S GPU).
Had InceptionTime been constrained to the same computational resources as MPSTime (i.e., a single CPU core), the training time would increase substantially.
Although 1-NN-DTW remains the fastest to train ($9.90 \times 10^{-3}$\,s), as shown in Fig.~\ref{fig:classification-barplots} of the main text, it delivered markedly lower classification accuracy on the ECG dataset relative to all other algorithms.
Taken together, our preliminary results suggest a promising speed--accuracy trade-off for MPSTime.
\begin{table}[H]
    \centering
    \caption{\textbf{Training times (mean ± std.\ dev.) on the ECG dataset.}
    Training time (in seconds) across 30 resample folds of the ECG dataset is shown for the classifiers considered in our benchmark.
    Note that computation time results are not fairly comparable for HIVE-COTE V2.0 (HC2) due to its algorithmic formulation, and so results are not reported.
    }
    \begin{tabular}{
        l
        S[table-format=1.2e+1]
        S[table-format=1.2e+1]
        c}
        \toprule
        \textbf{Classifier} & \textbf{Mean (s)} & \textbf{Std. Dev. (s)} & \textbf{Device} \\
        \hline
        1-NN-DTW      & 9.90e-3 & 2.81e-4 & CPU\textsuperscript{\S} \\
        MPSTime       & 1.90e+01  & 2.39e-1 & CPU\textsuperscript{\S} \\
        InceptionTime & 1.72e+03  & 1.28e+01 & GPU\textsuperscript{\textdaggerdbl} \\
        HC2 & \multicolumn{1}{c}{--} & \multicolumn{1}{c}{--} & \multicolumn{1}{c}{--} \\
    \end{tabular}
    \\
    \footnotesize \textsuperscript{\S} All CPU-based timings were performed on a single core of an INTEL(R) XEON(R) GOLD 6548N CPU.\\
    \textsuperscript{\textdaggerdbl} GPU timings used an entire NVIDIA L40S GPU.
    \label{tab:ecg-fit-time}
\end{table}

\section{Time-series imputation benchmarks}
\label{appenidx:imputation-baselines}
In the main text, we compare our imputation algorithm with several state-of-the-art methods, each representing different approaches to time-series imputation: (i) classical (1-NNI); (ii) diffusion model (CSDI) \cite{Tashiro2021:CSDIConditionalScorebased}; (iii) recurrent neural network (BRITS-I) \cite{caoBRITSBidirectionalRecurrent2018}; and (iv) matrix completion method (CDRec) \cite{khayatiMemoryefficientCentroidDecomposition2014}.
Specifically:
\begin{itemize}[noitemsep]
    \item \textbf{1-NNI}~\cite{Beretta2016:NearestNeighborImputation}: A special case of $k$-NNI (k-Nearest Neighbors Imputation), where $k = 1$.
    1-NNI substitutes missing data points in unseen time series with those from the Euclidean nearest neighbor time series in the training set.
    \item \textbf{CSDI}~\cite{Tashiro2021:CSDIConditionalScorebased}: CSDI (Conditional Score-based Diffusion Imputation) is a generative approach to imputation which uses a neural network-based diffusion model to infer new data points that are consistent with the underlying data distribution.
    \item \textbf{BRITS-I}~\cite{caoBRITSBidirectionalRecurrent2018}: BRITS-I (Bidirectional recurrent imputation for time series) is an adaptation of the BRITS algorithm to handle univariate time-series imputation.
    BRITS-I uses a recurrent neural network (RNN) architecture to learn a recurrent dynamical system from the observed data, and uses the learned recurrent dynamics to impute missing values in time series.
    \item \textbf{CDRec}~\cite{khayatiMemoryefficientCentroidDecomposition2014} is a matrix-based technique for memory efficient centroid decomposition (an approximation of SVD) of long time series.
\end{itemize}
For our benchmarking investigations in Sec.~\ref{sec:real-world-imputation-results}, we used off-the-shelf implementations of the aforementioned algorithms which are readily available in the Python \texttt{PyPOTS (v0.8)} package \cite{du2023pypots}.
Hyperparameters for each model were set to the default values provided with the implementation.

\bibliography{qml_refs, refs_extra}

\end{document}